\documentclass[11pt]{article}

\usepackage[final]{acl}

\usepackage{times}
\usepackage{latexsym}
\usepackage{enumitem}
\usepackage{booktabs}
\usepackage{amsmath}
\usepackage{amssymb}
\usepackage[most]{tcolorbox}
\usepackage{multirow}
\usepackage{makecell}
\usepackage{siunitx}
\usepackage{adjustbox}
\usepackage{fontawesome5}

\usepackage[T1]{fontenc}

\usepackage[utf8]{inputenc}

\usepackage{microtype}

\usepackage{inconsolata}

\usepackage{graphicx}

\definecolor{retrievalicon}{HTML}{9467BD}
\NewDocumentCommand\emojiretrieve{}{\,{\scriptsize\textcolor{retrievalicon}{\faDatabase}}}

 \title{Are you Synthesizing or Recalling? \\ Evaluating LLMs on Algorithmic Code Retrieval}

\author{Nickil Maveli \quad Antonio Vergari\textsuperscript{\emojiretrieve} \quad Shay B. Cohen\textsuperscript{\emojiretrieve} \\
  School of Informatics, University of Edinburgh \\
  10 Crichton Street, Edinburgh, EH8 9AB \\
  \texttt{\{nickil.maveli,avergari,scohen\}@ed.ac.uk}}

\newcommand{\datasetname}{\textsc{AlgoREval}}

\begin{document}
\maketitle

\begin{abstract}
Large language models (LLMs) have demonstrated strong performance in code generation, where success depends on both recalling relevant algorithmic knowledge and reasoning about how to apply it. %
However, existing LLM pipelines are opaque, with no explicit separation between these two components. 
We argue that for well-known algorithms whose canonical implementations are widely accessible in pretraining corpora, code generation is better measured as \textit{parametric code retrieval}: reproducing a named algorithm from internalised knowledge rather than synthesizing a novel one. 
We introduce \datasetname{}, a benchmark of 599 problems spanning classical 77 algorithms across 14 domains, 7 programming languages, and 4 graph-input representations to evaluate this capability in isolation, and assess 15 models (7B--34B parameters) in a zero-shot setting. 
We find substantial variation in retrieval accuracy across languages and input representations, even for widely documented algorithms and show that prompt augmentation with retrieved code snippets or structured algorithmic hints improve accuracy on complex algorithms, while SFT achieves broader language gains and GRPO achieves larger per-language gains on specific languages.
Together, our results establish parametric code retrieval as a distinct, measurable capability and caution against deploying AI-generated algorithmic code without systematic validation.\footnote{Code and dataset are available at \url{https://github.com/Nickil21/AlgoREval}.
{\emojiretrieve} denotes shared supervision.}
\end{abstract}

\section{Introduction}
\label{sec:intro}

\begin{figure}[t]
    \centering
    \includegraphics[width=1.\linewidth]{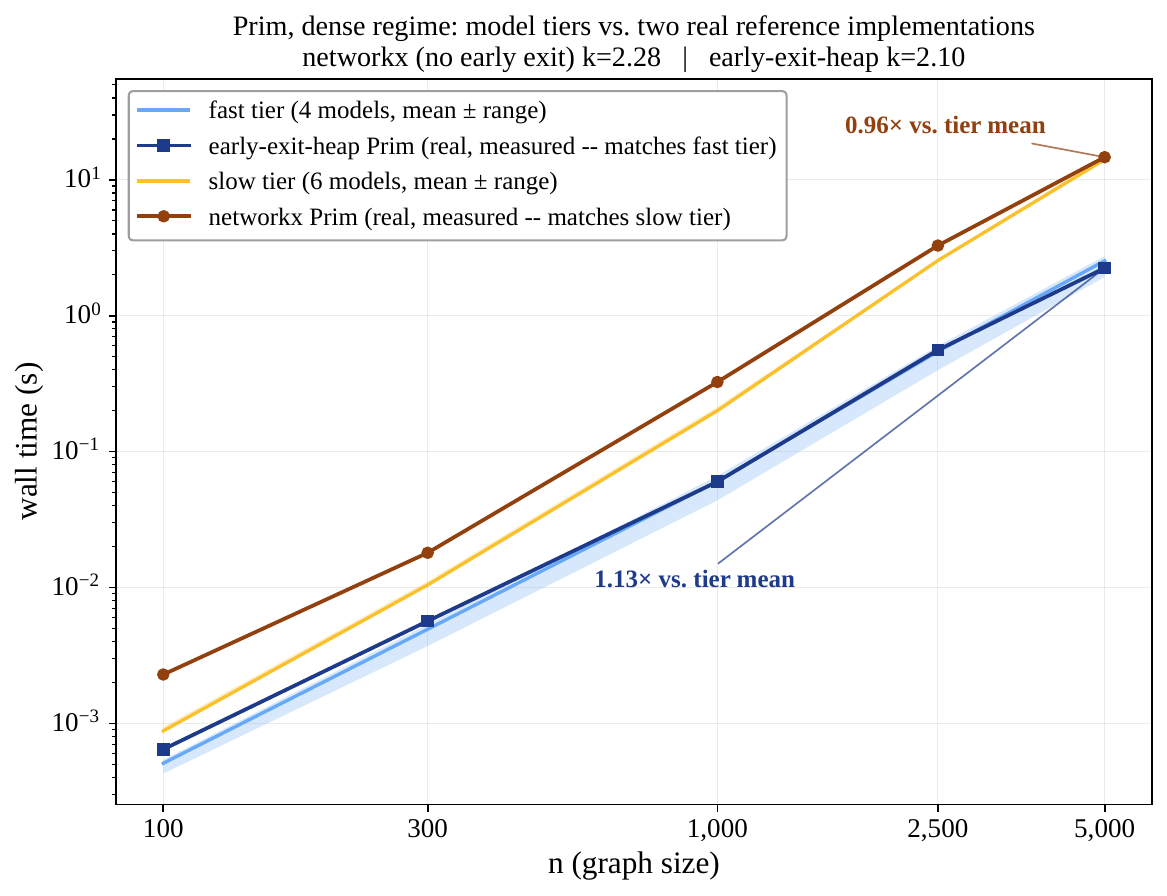}
    \caption{Ten LLMs implement Prim's MST algorithm, split into a fast tier (4 models) and a slow tier (6 models) by wall-clock scaling. We time two independently-written, correctness-verified reference implementations. Since both tiers land close to a real, legitimate pattern rather than one looking like an unexplained outlier, the evidence here is symmetric and consistent with both tiers reflecting retrieval of different existing implementations, not one retrieving and the other synthesizing from scratch.}
    \label{fig:placeholder}
\end{figure}


\begin{figure*}[t]
    \centering
    \begingroup
    \definecolor{BenchBg}{HTML}{DBEAFE}
    \definecolor{BenchBorder}{HTML}{60A5FA}
    \definecolor{PreBg}{HTML}{FAF5FF}
    \definecolor{PreBorder}{HTML}{C4B5FD}
    \definecolor{ModelBg}{HTML}{F8FAFC}
    \definecolor{OpenBg}{HTML}{DCFCE7}
    \definecolor{OpenBorder}{HTML}{22C55E}
    \definecolor{ClosedBg}{HTML}{FEF3C7}
    \definecolor{ClosedBorder}{HTML}{F59E0B}
    \definecolor{CodeBg}{HTML}{F6F8FB}
    \definecolor{CodeBgClosed}{HTML}{F2F5FA}
    \definecolor{CodeText}{HTML}{1F2937}
    \definecolor{CodeDim}{HTML}{64748B}
    \definecolor{CodeRule}{HTML}{D7DEE8}
    \definecolor{VsBlue}{HTML}{2563EB}
    \definecolor{VsTeal}{HTML}{0F766E}
    \definecolor{VsGreen}{HTML}{15803D}
    \definecolor{VsOrange}{HTML}{C2410C}
    \definecolor{VsPurple}{HTML}{7C3AED}
    \definecolor{VsGray}{HTML}{475569}
    \definecolor{EvalBg}{HTML}{FEF3C7}
    \definecolor{EvalBorder}{HTML}{F59E0B}
    \definecolor{PassBg}{HTML}{D1FAE5}
    \definecolor{PassBorder}{HTML}{10B981}
    \definecolor{DivBg}{HTML}{E0E7FF}
    \definecolor{DivBorder}{HTML}{6366F1}
    \definecolor{BadFail}{HTML}{FDA4AF}
    \definecolor{BadPass}{HTML}{86EFAC}
    \definecolor{GhBg}{HTML}{111827}
    \definecolor{SoBg}{HTML}{FFF1E6}
    \definecolor{SoBorder}{HTML}{F97316}
    \definecolor{RcBg}{HTML}{F3F4F6}
    \definecolor{RcBorder}{HTML}{9CA3AF}
    \definecolor{GfgBg}{HTML}{ECFDF5}
    \definecolor{GfgBorder}{HTML}{16A34A}
    \definecolor{DocBg}{HTML}{EFF6FF}
    \definecolor{DocBorder}{HTML}{3B82F6}
    \setlength{\fboxsep}{5pt}
    \setlength{\fboxrule}{0.8pt}

    \newcommand{\pill}[3]{{\setlength{\fboxsep}{1.2pt}\colorbox{#1}{\textcolor{#2}{\scriptsize\sffamily #3}}}}
    \newcommand{\logobox}[1]{\raisebox{-0.18em}{\setlength{\fboxsep}{1.2pt}\colorbox{white}{\includegraphics[height=1.1em,width=1.1em,keepaspectratio]{images/#1.png}}}}
    \newcommand{\langbadge}[5]{\logobox{#1}\hspace{0.35em}{\sffamily\bfseries #4}\hspace{0.22em}{\color{gray!45}$\cdot$}\hspace{0.18em}\pill{#2}{#3}{#5}}
    \newcommand{\llmlogo}{\raisebox{-0.28em}{%
      \begin{tikzpicture}[x=1em,y=1em,baseline]
        \draw[PreBorder,line width=0.35pt] (0.10,0.15) -- (0.45,0.50) -- (0.80,0.15);
        \draw[PreBorder,line width=0.35pt] (0.10,0.85) -- (0.45,0.50) -- (0.80,0.85);
        \fill[PreBorder] (0.10,0.15) circle (0.06);
        \fill[PreBorder] (0.10,0.85) circle (0.06);
        \fill[PreBorder] (0.45,0.50) circle (0.07);
        \fill[PreBorder] (0.80,0.15) circle (0.06);
        \fill[PreBorder] (0.80,0.85) circle (0.06);
      \end{tikzpicture}%
    }}
    \newcommand{\stagecard}[3]{\fcolorbox{#1}{#2}{\parbox[t]{0.5\linewidth}{\centering #3}}}
    \newcommand{\widecard}[3]{\fcolorbox{#1}{#2}{\parbox[t]{1.\linewidth}{\raggedright #3}}}
    \newcommand{\sourceicon}[4]{\raisebox{-0.15em}{\setlength{\fboxsep}{1.4pt}\fcolorbox{#2}{#1}{\textcolor{#3}{\scriptsize\sffamily\bfseries #4}}}}
    \newcommand{\sourcebadge}[5]{\sourceicon{#1}{#2}{#3}{#4}\hspace{0.28em}{\scriptsize\sffamily #5}}
    \newcommand{\sourcepng}[2]{\raisebox{-0.12em}{\includegraphics[height=0.56em]{images/#1}}\hspace{0.16em}{\scriptsize\sffamily #2}}
    \newcommand{\carddivider}{\vspace{1pt}{\color{CodeRule}\rule{\linewidth}{0.35pt}}\vspace{2pt}}
    \newcommand{\resultok}[1]{{\setlength{\fboxsep}{1.4pt}\colorbox{BadPass!18}{\textcolor{VsGreen}{\scriptsize\sffamily\bfseries PASS}}}\hspace{0.35em}{\scriptsize\sffamily\color{CodeDim}#1}}
    \newcommand{\resultbad}[1]{{\setlength{\fboxsep}{1.4pt}\colorbox{BadFail!16}{\textcolor{VsOrange}{\scriptsize\sffamily\bfseries FAIL}}}\hspace{0.35em}{\scriptsize\sffamily\color{CodeDim}#1}}
    \newtcblisting{predcard}[3]{%
      enhanced,
      width=\linewidth,
      colback=#2,
      colframe=#1,
      boxrule=0.8pt,
      arc=0pt,
      outer arc=0pt,
      left=4pt,
      right=4pt,
      top=4pt,
      bottom=4pt,
      listing only,
      breakable,
      listing engine=listings,
      listing options={
        basicstyle=\ttfamily\fontsize{4.2}{4.2}\selectfont\color{CodeText},
        breaklines=true,
        breakatwhitespace=false,
        columns=fullflexible,
        keepspaces=true,
        tabsize=2,
        showstringspaces=false,
        lineskip=-1pt,
        aboveskip=0pt,
        belowskip=0pt,
      },
      title={#3},
      coltitle=black,
      fonttitle=\scriptsize\sffamily\bfseries,
      attach boxed title to top left={xshift=0pt,yshift*=-\tcboxedtitleheight/2},
      boxed title style={
        boxrule=0pt,
        arc=0pt,
        outer arc=0pt,
        left=0pt,
        right=0pt,
        top=0pt,
        bottom=0pt,
        colback=#2,
      },
    }

    \noindent
    \begin{minipage}[t]{0.60\linewidth}
    \vspace{0pt}
    \fcolorbox{BenchBorder}{BenchBg}{\parbox[t]{1.\linewidth}{%
      {\small\sffamily\bfseries Benchmark Problem}\\[-3pt]
      {\small\textit{Bridge Finding}}\\[-3pt]
      {\tiny\selectfont\ttfamily\noindent
      Implement Tarjan's bridge-finding algorithm
      in JavaScript $|$ Java $|$ Python $|$ Rust.\\[-6pt]
      Implement findBridges(V, edges) where V is
      the vertex count (0-based) and edges is a\\[-6pt]
      list of [u,v] undirected pairs. For V=5,
      edges=[[0,1],[0,2],[1,2],[0,3],[3,4]],\\[-6pt]
      print each bridge as ``u v'' on its own
      line, sorted in ascending order.\\[-6pt]
      Expected output: ``0 3'' then ``3 4''.
      Runnable as a script; standard library only.}}}
    \end{minipage}\hfill%
    \begin{minipage}[t]{0.35\linewidth}
    \vspace{0pt}
    \centering
    \resizebox{\linewidth}{!}{%
    \begin{tikzpicture}[font=\sffamily\footnotesize]
      \definecolor{DiagTextBg}{HTML}{DDF4EC}
      \definecolor{DiagTextBorder}{HTML}{2A9D8F}
      \definecolor{DiagCodeBg}{HTML}{E3EEF9}
      \definecolor{DiagCodeBorder}{HTML}{3A7BC8}
      \definecolor{DiagPreBg}{HTML}{F9EEDB}
      \definecolor{DiagPreBorder}{HTML}{B7791F}
      \definecolor{DiagLlmBg}{HTML}{E9E8FB}
      \definecolor{DiagLlmBorder}{HTML}{6C63D9}
      \definecolor{DiagLayerBg}{HTML}{F7E7E1}
      \definecolor{DiagLayerBorder}{HTML}{C96C3A}
      \definecolor{DiagEmbedBg}{HTML}{F4F0E6}
      \definecolor{DiagEmbedBorder}{HTML}{8C8579}

      \node[anchor=west] at (0.,7.0) {\colorbox{DiagTextBg}{\phantom{xx}}~~Text};
      \node[anchor=west] at (0.,6.2) {\colorbox{DiagCodeBg}{\phantom{xx}}~~Code};

      \foreach \name/\x/\lbl in {tb/2.6/Books,tw/4.6/Web,twiki/6.5/Wikipedia,tp/8.7/Papers,td/10.8/Docs} {
        \node[
          draw=DiagTextBorder,
          rounded corners=0.15cm,
          fill=DiagTextBg,
          text width=1.5cm,
          align=center,
          font=\footnotesize,
          minimum height=0.5cm
        ] (\name) at (\x,7.0) {\lbl};
      }
      \foreach \name/\x/\lbl in {cg/3.0/GitHub,cso/5.2/Stack Overflow,crc/7.7/Rosetta Code,cgfg/10.0/GeeksforGeeks} {
        \node[
          draw=DiagCodeBorder,
          rounded corners=0.15cm,
          fill=DiagCodeBg,
          text width=2cm,
          align=center,
          font=\footnotesize,
          minimum height=0.5cm
        ] (\name) at (\x,6.2) {\lbl};
      }

      \node[draw=DiagPreBorder, rounded corners=0.18cm, fill=DiagPreBg, minimum width=3.1cm, minimum height=0.8cm] (pre) at (7.0,4.5) {Pre-training (offline)};

      \foreach \src in {tb,tw,twiki,tp,td,cg,cso,crc,cgfg} {
        \draw[gray!60] (\src.south) -- (pre.north);
      }

      \node[draw=DiagLlmBorder, rounded corners=0.35cm, fill=DiagLlmBg, minimum width=6.7cm, minimum height=3.6cm] (llm) at (7.0,2) {};
      \node[anchor=north,font=\large] at (llm.north) {LLM};

      \node[draw=DiagLayerBorder, rounded corners=0.14cm, fill=DiagLayerBg, minimum width=5.4cm, minimum height=0.5cm] at (7.0,2.9) {Layer N};
      \node[draw=DiagLayerBorder, rounded corners=0.14cm, fill=DiagLayerBg, minimum width=5.4cm, minimum height=0.5cm] at (7.0,2.15) {Layer \ldots};
      \node[draw=DiagLayerBorder, rounded corners=0.14cm, fill=DiagLayerBg, minimum width=5.4cm, minimum height=0.5cm] at (7.0,1.40) {Layer 1};
      \node[draw=DiagEmbedBorder, rounded corners=0.14cm, fill=DiagEmbedBg, minimum width=5.4cm, minimum height=0.42cm] at (7.0,0.72) {Embedding};
      \node[anchor=south,font=\footnotesize\itshape] at (7.0,0.10) {Parameters $\theta$};

      \draw[->,gray!70,line width=0.5pt] (pre.south) -- (llm.north);
      \draw[->,gray!70,line width=0.5pt] (3.1,2) -- (llm.west);
      \node[anchor=east] at (3.1,2) {Query};
      \draw[->,gray!70,line width=0.5pt] (llm.east) -- (10.7,2);
      \node[anchor=west] at (10.7,2) {Response};
    \end{tikzpicture}}%
    \end{minipage}

    \noindent
    \begin{minipage}[t]{0.244\linewidth}
    \begin{predcard}{OpenBorder}{CodeBg}{\langbadge{python}{blue!60!black}{yellow}{Yi-Coder-9B}{Python}}
def findBridges(V, edges):
    adj = [[] for _ in range(V)]
    disc = low = [-1] * V
    # ... for u,v in edges: adj[u]+=...
    def dfs(u):
        disc[u] = low[u] = self.time
        self.time += 1
        # ... update low, collect bridges ...
    self.time = 0
    bridges = []
    for u in range(V):
        if not visited[u]: dfs(u)
    return bridges
\end{predcard}
    \centering\resultbad{runtime error}
    \end{minipage}\hspace{0.008\linewidth}%
    \begin{minipage}[t]{0.244\linewidth}
    \begin{predcard}{OpenBorder}{CodeBg}{\langbadge{rust}{gray!25}{black}{CodeLlama-34B}{Rust}}
use std::collections::HashMap;

fn find_bridges(v: usize, ...) {
    let mut graph: HashMap<usize,Vec<usize>>
        = HashMap::new();
    // ... build graph, init arrays ...
    let mut br = Vec::new();
    fn dfs(g: &HashMap<usize,Vec<usize>>,
           v: usize, /* ... */) { /* ... */ }
    for i in 0..v { if !vis[i] {
        dfs(graph, i, /* ... */); } }
    br
}
\end{predcard}
    \centering\resultbad{compile error}
    \end{minipage}\hspace{0.008\linewidth}%
    \begin{minipage}[t]{0.244\linewidth}
    \begin{predcard}{ClosedBorder}{CodeBgClosed}{\langbadge{java}{orange!80!black}{white}{gpt-4o-mini}{Java}}
import java.util.*;

public class Main {
    private int time = 0;
    List<int[]> findBridges(int V, ...) {
        // ... build graph g, init arrays ...
        List<int[]> bridges=new ArrayList<>();
        for (int i=0;i<V;i++)
            if (!vis[i]) dfs(i,-1,...,bridges);
        return bridges;
    }
    void dfs(int u, int par, ...) { /* ... */ }
}
\end{predcard}
    \centering\resultbad{wrong output order}
    \end{minipage}\hspace{0.008\linewidth}%
    \begin{minipage}[t]{0.244\linewidth}
    \begin{predcard}{OpenBorder}{CodeBg}{\langbadge{javascript}{yellow!80!black}{black}{QwQ-32B}{JavaScript}}
function findBridges(V, edges) {
    const bridges = [];
    // ... build adj, init disc/low/timer ...
    function dfs(at, parent) {
        disc[at]=low[at]=timer++;
        // ... update low values ...
        if (low[to]>disc[at])
            bridges.push([Math.min(at,to),
                          Math.max(at,to)]);
    }
    bridges.sort((a,b)=>a[0]-b[0]||a[1]-b[1]);
    return bridges;
}
\end{predcard}
    \centering\resultok{correct output}
    \end{minipage}
    \endgroup
    \caption{\textbf{Parametric code retrieval evaluation pipeline.} LLMs generate solutions zero-shot, drawing solely on algorithmic knowledge encoded in their pretraining weights. The four panels show four models (three open-weight, one closed API) on the same Bridge Finding problem, each in a different language: Yi-Coder-9B (Python) uses \texttt{self.time} in a plain function with no class, causing a \textit{runtime error}; CodeLlama-34B (Rust) passes \texttt{graph} by value where a reference is required, producing a \textit{compile error}; gpt-4o-mini (Java, closed API) finds the correct bridges but omits the sort step, yielding \textit{wrong output order}; QwQ-32B (JavaScript) normalises, sorts, and passes.
    }
    \label{fig:pipeline}
\end{figure*}


When an LLM writes Dijkstra's algorithm, is it \textit{synthesizing}, \textit{retrieving}, or \textit{reciting}? 
\textit{Synthesizing} means reasoning from the problem statement to a solution, deriving logic the model has not seen in the exact form during pre-training~\citep{chen2021codex,DBLP:journals/corr/abs-2108-07732}.
\textit{Retrieving} means reconstructing a correct implementation from internalised parametric knowledge. The model has encoded the algorithm's structure, invariants, and control flow in its weights and can express it correctly even in forms it has not seen verbatim~\citep{roberts-etal-2020-much,gekhman2026thinkingrecallreasoningunlocks}.
\textit{Reciting} means replaying a near-verbatim token sequence cached from a specific training example: correct output is a side-effect of having stored that string, not of understanding what it computes~\citep{carlini2023quantifying,huang-etal-2024-demystifying}.
Synthesis, retrieval, and recitation are distinct capabilities with different failure modes and different implications for reliability: synthesis may fail on novel problems; retrieval degrades when surface form shifts away from the training distribution; recitation breaks the moment the expected transcription is absent.
Prior work conflates all three under the label of \textit{code generation}. 

We focus on the ``retrieval'' aspect.
The distinction has real stakes: an estimated 72\% of developers now use AI coding tools on a daily basis, and AI currently generates 42\% of all committed code.\footnote{\url{https://www.sonarsource.com/state-of-code-developer-survey-report.pdf}} 
Yet, 66\% report their top frustration is code that is almost right, but not quite, which precisely describes retrieval failure: the model knows the algorithm but gets a detail wrong~\cite{zhu2025llmagentsfaillearn}.
In agentic coding pipelines, where models plan, generate, and commit code without per-step human review, a retrieval failure in a graph algorithm may not surface until a production incident exposes the wrong output being computed silently~\citep{sapkota2025vibecodingvsagentic,NEURIPS2025_b1041e52}. Knowing where and how retrieval breaks is the minimum information needed to decide when AI-generated algorithmic code can be trusted without verification.

Most code generation research frames the task as \textit{synthesis}: given a problem statement, construct a solution. This is appropriate for novel problems, but for well-known algorithms whose canonical implementations appear across thousands of open-source repositories and almost certainly in the pretraining corpora of every model we evaluate. The model need not invent anything; it must \textit{retrieve}. Conflating the two obscures a distinct failure mode: a model may demonstrate impressive synthesis on novel tasks while quietly failing to reproduce even the most widely documented algorithms, or vice-versa~\citep{riddell-etal-2024-quantifying,matton-etal-2024-leakage}. We isolate this as \textit{parametric code retrieval}: reproducing a correct implementation of a
named algorithm in a specified language from parametric memory alone. Code constitutes roughly 13\% of the Dolma corpus (411\,B of 3.1\,T tokens)~\citep{kocetkov2022stack3tbpermissively,dolma}, with JavaScript (486\,GB), Java (271\,GB), and C (222\,GB) accounting for over half by volume, which are precisely the language families our benchmark targets.

\textbf{Contributions.} We introduce \datasetname{}, a benchmark of 599 problems spanning 77 algorithms, 7 programming
languages, and 4 graph-input representations, and evaluate fifteen models zero-shot. We then ask whether retrieval failures can be remediated: first via prompt augmentation with retrieved code snippets or structured algorithmic hints, and then via GRPO (Group Relative Policy Optimization) fine-tuning with execution-based rewards compared against a SFT (Supervised fine-tuning) baseline. Our main findings are threefold. First, retrieval accuracy varies widely across models, and the variance is governed by surface form (programming language and input representation) and not model size or code specialisation. The dominant failure mode in Python and JavaScript is wrong output: runnable code that silently returns the wrong answer. Second, prompt augmentation does not reliably improve retrieval in aggregate, and actively hurts when parametric knowledge is already present; executable snippets are more effective than abstract hints only on problems where the model's parametric trace is already thin. Third, GRPO fine-tuning with execution-based rewards converges faster than SFT and closes language gaps that binary supervision cannot address, particularly in Rust where partial-credit signals from compile failures carry information that correct/incorrect labels discard.


\section{Task Formulation}

We define \textit{parametric code retrieval} as the task of recovering a correct, executable implementation of a named algorithm from an LLM's parametric memory, in a specified programming language and input representation, without access to external documents or in-context examples.

\paragraph{Problem instances.}
Let $\mathcal{A}$ denote a set of algorithm names, $\mathcal{L} = \{\text{Python, C\texttt{++}, Java, C, JavaScript, Rust, Go}\}$ a set of target languages, and $\mathcal{R}_a$ the set of input-representation formats applicable to algorithm $a$ (e.g., edge list, adjacency list, adjacency matrix for graph algorithms; a single canonical format for all others). Each problem instance $p \in \mathcal{P}$ is a tuple
\[
  p = (a,\; \ell,\; r,\; \sigma_p,\; t_p,\; y_p),
\]
where $\sigma_p$ is the required function or class signature, $t_p$ is a fixed test input, and $y_p$ is the expected standard output produced by executing a reference implementation on $t_p$. Each instance is additionally tagged with a domain label $d \in \mathcal{D}$ used for stratified reporting but not part of the prompt or scoring. Multiple instances may share the same $(a, \ell)$ while differing only in $r$, the different input representations; these probe representational robustness of the same underlying algorithm without changing its identity.

\paragraph{Prompt structure.}
The prompt $x_p$ names the algorithm $a$, specifies the target language $\ell$ and input representation $r$, states the required signature $\sigma_p$, and embeds the concrete test input $t_p$. No reference implementation, pseudocode, or chain-of-thought scaffolding is provided. This zero-shot constraint ensures that any correct solution must originate from the model's parametric memory rather than from in-context retrieval.

\paragraph{Scoring.}
A model $m$ generates code $c_m(p)$ from $x_p$. Executing $c_m(p)$ in a sandboxed subprocess yields output $\hat{y}_m(p)$, and retrieval succeeds when $\hat{y}_m(p) = y_p$ (stdout match after whitespace normalisation). We use the pass rate~\citep{chen2021codex} of model $m$ over subset $S \subseteq \mathcal{P}$ as
\[
  \mathrm{Pass}(m,\, S) \;=\; \frac{1}{|S|} \sum\nolimits_{p \in S} \mathbf{1}\!\left[\hat{y}_m(p) = y_p\right],
\]
and report $\mathrm{Pass}(m, \mathcal{P})$ overall, stratified by language $\ell$ and domain $d$ across all instances, and by representation $r$ within the graph-algorithm subset $\mathcal{P}_{\text{graph}} \subset \mathcal{P}$ where multiple representations exist. Failures fall into four categories: \textit{static failure} (code fails to parse or compile before execution), \textit{runtime error} (code raises an exception during execution), \textit{timeout} (execution exceeds the time budget), and \textit{wrong output} ($\hat{y}_m(p) \neq y_p$ despite clean execution). Wrong-output failures are the most diagnostic: they show the model produced algorithmically plausible but incorrect logic.



\section{Dataset Construction}


We build the benchmark with 107 Python problems drawn from RosettaCode\footnote{\url{https://rosettacode.org/wiki/Rosetta_Code}} and GeeksforGeeks\footnote{\url{https://www.geeksforgeeks.org}}, covering 77 distinct algorithms across 14 domains: graph algorithms, sorting, dynamic programming, string algorithms, number theory, image processing, data structures, computational geometry, numerical methods, compression, cryptography, backtracking, and parsing.
Some algorithms contribute to more than one problem, e.g., a weighted and an unweighted variant, accounting for the gap between 107 problems and 77 algorithms.
We select algorithms that are extensively documented across open-source repositories and therefore almost certainly present in any large pretraining corpus.
This is a deliberate design choice: we are not testing generalisation to unseen problems, but the fidelity of recall for algorithms a model has already been exposed to.
Ground truth for each problem is established by executing a reference implementation on the fixed test input $t_p$ and recording its standard output as $y_p$; a model passes if and only if its generated code reproduces $y_p$ exactly after whitespace normalisation.
Each algorithm is assigned to exactly one primary domain based on its canonical problem family, not all techniques it may employ: an expression parser is labeled \textit{parsing} rather than dynamic programming, even if it uses memoisation internally.

Each Python problem is adapted to six additional languages (C\texttt{++}, Java, C, JavaScript, Rust, and Go) by applying three transformations: (1)~replacing the target language name throughout the prompt, (2)~substituting the embedded Python test block with an explicit expected-output specification, and (3)~rewriting Python-specific entry-point idioms to language-appropriate forms (e.g., \texttt{int main()} for C/C\texttt{++}, \texttt{fn main()} for Rust).
This yields a multilingual benchmark in which the same algorithmic task is posed identically across all 7 languages, isolating language-level retrieval differences from differences in problem framing.

Parametric retrieval is interface-sensitive: a model that correctly implements Dijkstra’s algorithm given an edge list may fail when the same graph is presented as an adjacency matrix, even though the underlying algorithm is unchanged.
To probe this, we augment the 24 graph-algorithm problems with variants across four input representations, yielding 228 graph-algorithm entries in total:

\begin{itemize}[topsep=0pt,noitemsep,parsep=0pt,leftmargin=*]
  \item \textbf{Edge list} (\texttt{edge\_list}): the default; \texttt{solve(n, edges, \ldots)} where \texttt{edges} is a list of \texttt{(u,v)} or \texttt{(u,v,w)} tuples.
  \item \textbf{Adjacency list} (\texttt{adj\_list}): \texttt{solve(adj, \ldots)} where \texttt{adj[u]} is a list of neighbours or (neighbour, weight) pairs; vertex count is implicit as \texttt{len(adj)}.
  \item \textbf{Adjacency matrix} (\texttt{adj\_matrix}): \texttt{solve(M, \ldots)} where \texttt{M[u][v]} encodes edge weight ($\infty$ for absent edges; $1/0$ for unweighted graphs).
  \item \textbf{Incidence matrix} (\texttt{incidence\_matrix}): an $(n{\times}m)$ vertex-by-edge matrix used for selected bipartite-matching problems.
\end{itemize}

\begin{table}[t]
\centering
\resizebox{0.7\linewidth}{!}{%
\begin{tabular}{lrr}
\toprule
\textbf{Domain} & \textbf{Problem Types} & \textbf{Entries} \\
\midrule
Graph algorithms        & 24 & 228 \\
Image processing        & 10 & 70  \\
Data structures         & 5  & 35  \\
Computational geometry  & 4  & 28  \\
Sorting algorithms      & 4  & 28  \\
Dynamic programming     & 4  & 28  \\
Numerical methods       & 4  & 28  \\
Backtracking            & 3  & 21  \\
Cryptography            & 3  & 21  \\
String algorithms       & 3  & 21  \\
Number theory           & 3  & 21  \\
Parsing                 & 2  & 14  \\
Compression             & 2  & 14  \\
Other                   & 6  & 42  \\
\midrule
\textbf{Total}          & \textbf{77} & \textbf{599} \\
\bottomrule
\end{tabular}
}
\caption{\textbf{Dataset breakdown by domain.} Each problem type is instantiated across 7 languages. The graph-algorithm problems additionally include up to 4 input-representation variants.
}
\label{tab:dataset-stats}
\end{table}

Representation coverage varies by algorithm: adjacency list is available for 20 of the 24 graph algorithms, adjacency matrix for 10, and incidence matrix for 2.
The remaining 371 non-graph entries use edge list as their sole representation.
The final benchmark contains 599 entries spanning 7 programming languages and 4 input representations. Table~\ref{tab:dataset-stats} provides a full breakdown by domain and language.




\section{Related Work}
\subsection{Parametric Memory in LLMs.}
Pretraining encodes factual and procedural knowledge directly into model weights as
\textit{parametric memory}~\citep{petroni-etal-2019-language}, which can be queried at inference time without external retrieval. Probing studies on factual QA show that recall is reliable when inference-time surface forms closely match training data, but degrades sharply under paraphrase or format change~\citep{roberts-etal-2020-much,gekhman2026thinkingrecallreasoningunlocks}. We study the same degradation in code: the surface shift is a change in programming language or input representation rather than a change in natural-language phrasing, and the failure mode is wrong output rather than a factual error.

\subsection{Code Generation Benchmarks}
The dominant paradigm in code generation research treats the task as synthesis: given a
natural-language specification, produce a correct program.
HumanEval~\citep{chen2021codex} and MBPP~\citep{austin2021programsynthesislargelanguage} operationalise this with short functional puzzles. APPS~\citep{hendrycksapps2021} scales difficulty to competitive programming. LiveCodeBench~\citep{jain2024livecodebenchholisticcontaminationfree} reduces contamination by time-gating problems after model training cutoffs.
All of these treat training-data overlap as a confound to eliminate. 
However, \citet{riddell-etal-2024-quantifying} found that only 3.6--20.8\% of their reference solutions turn up almost word-for-word in common pretraining corpora such as the Pile~\citep{gao2020pile800gbdatasetdiverse} and the Stack~\citep{lozhkov2024starcoder2stackv2}. Either way, these benchmarks do not probe whether the model has seen the underlying algorithm before and can reproduce it correctly in a given language and input format. That framing is appropriate for novel problems, but not for algorithms that are widely documented, where correct behavior is explained by \textit{retrieval}: reconstructing a correct implementation from internalised parametric knowledge, expressing the algorithm's structure, invariants, and control flow correctly even in forms not seen verbatim.
We take the opposite stance: overlap is the \textit{precondition} for our task, and the question worth asking is not whether a model can produce code but whether it faithfully reproduces correct implementations of algorithms it has already encountered. Crucially, because existing benchmarks do not control for training exposure, a model's pass rate on a well-known algorithm conflates two fundamentally different capabilities: \textit{retrieval}, i.e., recalling a correct implementation from parametric memory, and \textit{understanding}, i.e., deriving a solution through reasoning~\citep{maveli2026llmscompressanddecompress}.

\subsection{Retrieval-augmented generation (RAG)}
RAG-based code generation assumes a parametric knowledge deficit and supplies retrieved context to fill it~\citep{wang-etal-2025-coderag}. The retrieved examples add retrieval noise, misaligned context, and surface-level code patterns that compete with, rather than complement, what the model already encodes parametrically. Models perform significantly better on the subset of benchmarks where similar solutions were seen during training, and are more likely to emit memorized training data when they have seen it multiple times~\citep{riddell-etal-2024-quantifying}. We therefore focus on parametric code recall: the model's capacity to retrieve correct algorithmic solutions from weights alone, without retrieval augmentation.


\section{Experimental Setup}
\subsection{Models}
Programming language distribution in LLM pretraining data varies dramatically by model type and is rarely disclosed at the per-language level. General-purpose models allocate a modest fraction to code: LLaMA-1~\citep{touvron2023llamaopenefficientfoundation} reserved only 4.5\% of its 1.4T tokens for GitHub content, while LLaMA-3~\citep{grattafiori2024llama3herdmodels} scaled this to approximately 17\% of 15T+ tokens. Code-centric models invest far more: DeepSeek-Coder~\citep{guo2024deepseekcoderlargelanguagemodel} allocates 87\% of its 2T tokens to code across 87 languages, and StarCoder2~\citep{lozhkov2024starcoder2stackv2} trains almost entirely on The Stack v2 (619 languages, \textasciitilde4T tokens). This variation matters for our benchmark: a model's pretraining code budget directly shapes which languages it has had sufficient exposure to for reliable parametric retrieval.

We evaluate 15 models spanning three axes of variation: (1) \textit{specialisation}: general-purpose instruction-tuned models versus code-specialised ones; (2) \textit{scale}: 7B to 34B parameters, and (3) \textit{inference regime}: standard decoding versus extended chain-of-thought reasoning. Full model identifiers, parameter counts, and licenses are listed in Appendix~\ref{sec:models}.

\subsection{Evaluation}
\paragraph{Execution harness.}
Each model-generated program is executed in an isolated subprocess with a 30-second wall-clock timeout.
Interpreted languages (Python, JavaScript) are run directly, whereas compiled languages (C, C\texttt{++}, Java, Rust, Go) go through a mandatory compilation step first.
A compilation failure is recorded immediately as a static failure without attempting execution.
Standard output is compared against the ground-truth string after whitespace normalisation; a submission passes if and only if the strings match exactly.
Failures are categorised into four mutually exclusive types: \textit{static failure} (parse or compile error), \textit{runtime error} (uncaught exception during execution), \textit{timeout} (execution exceeds 30\,s), and \textit{wrong output} (clean execution but incorrect stdout).

\paragraph{Metrics.}
Our primary metric is \textbf{pass rate}: the fraction of problem instances on which a model's generated code passes, reported overall and stratified by domain, language $\ell$, and input representation $r$.
Pass rate captures correctness but not consistency: two models can both pass a problem while producing structurally unrelated implementations.
To characterise solution diversity, we follow the AlgoSim evaluation framework~\citep{lee-etal-2025-diversely}: model predictions are pooled, grouped by problem label, and evaluated for pairwise lexical similarity across every pair of models that each produced code for the same problem.

With the benchmark and evaluation harness in place, we organise our experiments around three questions.
The first establishes where and how retrieval fails: without a clear failure map, any proposed remedy is unmotivated.
The second asks whether those failures can be patched at inference time through prompt augmentation, which requires no model changes but assumes that injecting external context into a prompt where parametric knowledge already exists is net-positive.
The third asks whether the gaps can instead be closed at the weight level through execution-based fine-tuning, separating the contribution of reward signal (GRPO) from the contribution of training on correct solutions alone (SFT).
In this paper, we mainly investigate the following research questions through experiments.
\begin{tcolorbox}[colback=gray!5,colframe=gray!40,left=6pt,right=6pt,top=4pt,bottom=4pt]
\textbf{RQ1.} How reliably do LLMs retrieve algorithmic code from parametric knowledge across languages and input representations?
\end{tcolorbox}
\begin{tcolorbox}[colback=gray!5,colframe=gray!40,left=6pt,right=6pt,top=4pt,bottom=4pt]
\textbf{RQ2.} Does augmenting LLM prompts with retrieved code snippets or structured algorithmic hints improve parametric code retrieval accuracy, and which source better compensates for failures on complex and esoteric algorithms?
\end{tcolorbox}
\begin{tcolorbox}[colback=gray!5,colframe=gray!40,left=6pt,right=6pt,top=4pt,bottom=4pt]
\textbf{RQ3.} Does GRPO fine-tuning with execution-based rewards reduce representation and language accuracy gaps relative to SFT on algorithmic code retrieval?
\end{tcolorbox}

\section{Analysis}
We report results for each research question in turn, moving from baseline measurement of retrieval fidelity (RQ1) through inference-time intervention via prompt augmentation (RQ2) to weight-level adaptation via fine-tuning (RQ3).
\subsection{RQ1: Parametric Code Retrieval Across Languages and Input Representations}



We evaluate all models in a zero-shot setting with no in-context examples, no chain-of-thought scaffolding, and no hints about the solution approach. The prompt template and generation hyperparameters are detailed in Appendices~\ref{sec:prompt-template} and~\ref{sec:hyperparams}.

\begin{figure}[t]
    \centering
    \includegraphics[width=\linewidth]{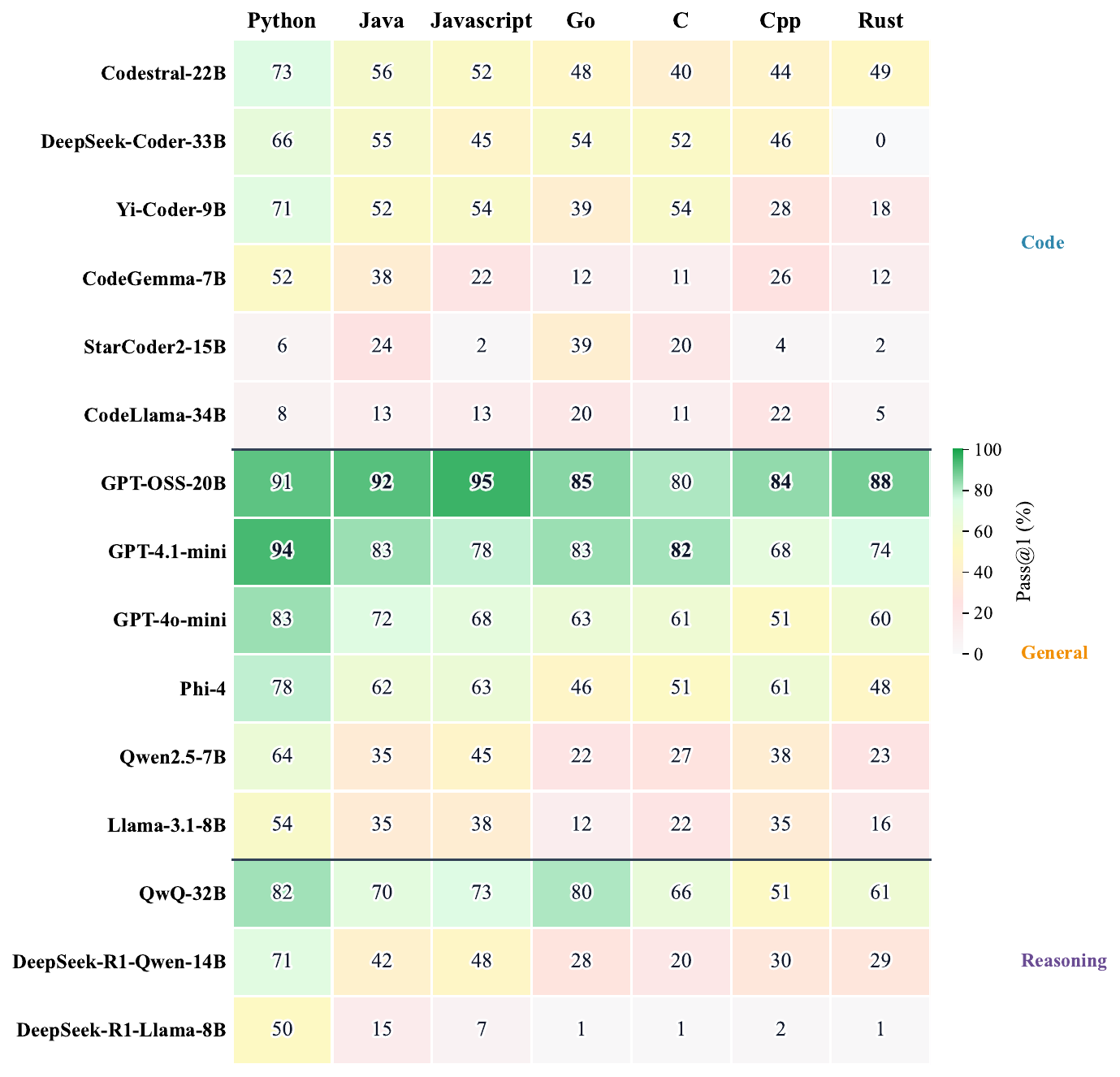}
    \caption{\textbf{Pass@1 heatmap by model and programming language.} Rows are grouped by model category (Code, General, Reasoning) and ordered by mean pass rate within each group.}
    \label{fig:pass-rate-lang}
\end{figure}

\begin{figure}[t]
    \centering
    \includegraphics[width=1.\linewidth]{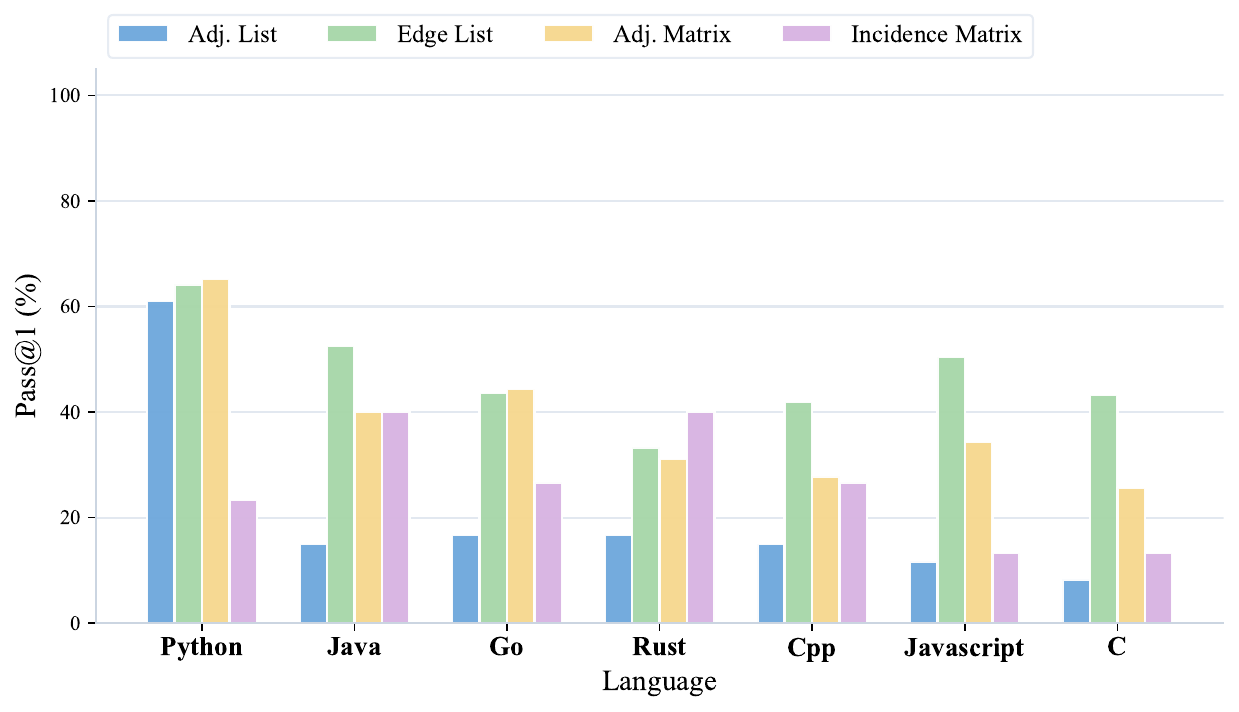}
    \caption{\textbf{Pass@1 by programming language, broken down by input representation and aggregated across all model predictions.} Each group compares edge-list, adjacency-list, adjacency-matrix, and incidence-matrix inputs within a language.}
    \label{fig:pass-rate-repr}
\end{figure}

The aggregate pass rate across all 15 models is 41.2\%, with substantial variation across languages and input representations but little correlation with model size or code specialisation (Table~\ref{tab:rq1-results}).
gpt-oss-20B achieves the highest overall pass rate despite having only 20B active parameters, outperforming larger models, including QwQ-32B (32B) and GPT-4o-mini.
Figure~\ref{fig:pass-rate-lang} breaks down pass rates by model and programming language: general-purpose models lead in most languages, while code-specialised models exhibit more uneven language coverage.
The dominant failure mode in Python and JavaScript is wrong output (runnable code that silently computes the incorrect answer), whereas compiled languages show a higher proportion of static failures (Figure~\ref{fig:error-analysis}).
Representation sensitivity is pronounced (Figure~\ref{fig:pass-rate-repr}): for the same model on the same algorithm, pass rates drop by over 10 percentage points when the input encoding shifts from edge list to adjacency matrix, consistent with the interface-sensitivity argument in Section~\ref{sec:intro}.

\begin{table}[ht]
\centering
\resizebox{0.8\linewidth}{!}{%
\begin{tabular}{llcrS[table-format=2.1]}
\toprule
\textbf{Category} & \textbf{Model} & \textbf{Weights} & \textbf{Params} & {\textbf{Pass@1 (\%)}} \\
\midrule
\multirow{6}{*}{Code}
  & Codestral-22B              & Open   & 22   & \textbf{52.6} \\
  & DeepSeek-Coder-33B         & Open   & 33   & 46.4 \\
  & Yi-Coder-9B                & Open   &  9   & 46.2 \\
  & CodeGemma-7B               & Open   &  7   & 25.9 \\
  & StarCoder2-15B             & Open   & 15   & 13.7 \\
  & CodeLlama-34B              & Open   & 34   & 12.9 \\
\midrule
\multirow{6}{*}{General}
  & gpt-oss-20B                & Open   & 20   & \textbf{88.0} \\
  & GPT-4.1-mini               & Closed & {--} & 80.8 \\
  & GPT-4o-mini                & Closed & {--} & 66.3 \\
  & Phi-4                      & Open   & 14   & 59.4 \\
  & Qwen2.5-7B                 & Open   &  7   & 37.4 \\
  & Llama-3.1-8B               & Open   &  8   & 31.4 \\
\midrule
\multirow{3}{*}{Reasoning}
  & QwQ-32B                    & Open   & 32   & \textbf{69.6} \\
  & DeepSeek-R1-Distill-Qwen-14B  & Open   & 14   & 39.6 \\
  & DeepSeek-R1-Distill-Llama-8B  & Open   &  8   & 12.9 \\
\bottomrule
\end{tabular}
}
\caption{\textbf{Zero-shot Pass@1 (\%) grouped by category and ranked within
each group by Pass@1}. Closed API models (GPT-4.1-mini, GPT-4o-mini) have undisclosed sizes.}
\label{tab:rq1-results}
\end{table}

\paragraph{Solution diversity.}
To test whether correct solutions reflect recitation of stored tokens or genuine reconstruction of algorithmic knowledge, we compute pairwise lexical similarity across all model predictions (2{,}625 pairs over 25 problems). The mean SequenceMatcher similarity is 0.21, and models produce syntactically distinct implementations even when solving the same problem correctly (EM=0 across all pairs). For semantic-level analysis we apply AlgoSim~\citep{lee-etal-2025-diversely}, which groups pairs by whether they share the same underlying algorithmic approach. Across 2{,}215 pairs spanning 37 algorithms, 80\% of pairs share the same underlying approach despite surface variation, while 20\% represent structurally distinct implementations of the same algorithm. This combination of high surface diversity, predominantly shared algorithmic structure, and zero verbatim matches is consistent with parametric retrieval: models converge on the same algorithm from memory but express it independently rather than replaying a cached token sequence. Figure~\ref{fig:consensus-recall} further shows that pass rate and consensus recall are correlated but not identical: some models produce correct outputs that deviate from the plurality answer, indicating independent retrieval rather than convergence on a single cached response. A breakdown of AlgoSim novel rate per model and language is provided in Figure~\ref{fig:diversity-model-lang}.

\subsection{RQ2: Prompt Augmentation via Retrieved Snippets and Structured Hints}
When parametric retrieval fails, a natural remedy is to inject external knowledge directly into the prompt. We test two distinct sources of such knowledge and ask whether either one can rescue failing models. The \textbf{+Hint} condition prepends language-agnostic pseudocode (up to 20 lines) generated offline by GPT-4o-mini, covering key data structures and asymptotic complexity. The \textbf{+Snippet} condition injects a language-matched reference implementation retrieved from GitHub (up to 80 lines, centred on the most relevant region), explicitly excluding RosettaCode and GeeksforGeeks to avoid overlap with our benchmark sources. Both conditions otherwise use identical prompts and generation hyperparameters to the zero-shot baseline. All 599 benchmark problems received a language-matched GitHub snippet. Full setup details and prompt examples appear in Appendices~\ref{sec:rq2-hyperparams} and~\ref{sec:augmentation-examples}.

\begin{table}[t]
\centering
\resizebox{0.8\linewidth}{!}{%
\newcommand{\up}[1]{{\color{teal}#1}}
\newcommand{\dn}[1]{{\color{red}#1}}
\newcommand{\nc}[1]{{\color{gray}#1}}
\begin{tabular}{lrrrrrr}
\toprule
\textbf{Model} & \textbf{Base} & \textbf{+Hint} & $\boldsymbol{\Delta}$\textbf{(H)} & \textbf{+Snip} & $\boldsymbol{\Delta}$\textbf{(S)} & \textbf{Better} \\
\midrule
StarCoder2-15B & 13.7 & 12.0 & \dn{-1.7} & 8.7 & \dn{-5.0} & H \\
DeepSeek-R1-Distill-Llama-8B & 12.9 & 11.2 & \dn{-1.7} & 12.4 & \dn{-0.5} & S \\
CodeGemma-7B & 25.9 & 26.0 & \up{+0.2} & 24.5 & \dn{-1.3} & H \\
CodeLlama-34B & 12.9 & 11.7 & \dn{-1.2} & 11.7 & \dn{-1.2} & tie \\
Llama-3.1-8B & 31.4 & 27.7 & \dn{-3.7} & 29.7 & \dn{-1.7} & S \\
Qwen2.5-7B & 37.4 & 37.6 & \up{+0.2} & 37.1 & \dn{-0.3} & H \\
Yi-Coder-9B & 46.2 & 46.2 & \nc{0.0} & 42.7 & \dn{-3.5} & H \\
DeepSeek-Coder-33B & 46.4 & 46.4 & \nc{0.0} & 47.4 & \up{+1.0} & S \\
DeepSeek-R1-Distill-Qwen-14B & 39.6 & 37.6 & \dn{-2.0} & 39.9 & \up{+0.3} & S \\
Phi-4 & 59.4 & 58.9 & \dn{-0.5} & 59.6 & \up{+0.2} & S \\
Codestral-22B & 52.6 & 50.4 & \dn{-2.2} & 52.8 & \up{+0.2} & S \\
QwQ-32B & 69.6 & 73.6 & \up{+4.0} & 74.6 & \up{+5.0} & S \\
gpt-oss-20B & 88.0 & 89.0 & \up{+1.0} & 90.2 & \up{+2.2} & S \\
\midrule
\textbf{Macro avg.} & 41.2 & 40.6 & \dn{-0.6} & 40.9 & \dn{-0.4} & H:4 S:8 \\
\bottomrule
\end{tabular}
}
\caption{\textbf{Pass rates (\%) under three conditions across 13 models on all 599 benchmark examples.} \textit{Base}: zero-shot baseline. \textit{+Hint}: GPT-4o-mini-generated pseudocode. \textit{+Snip}: language-matched GitHub snippet. \textbf{Better} indicates which augmentation condition outperforms the other per model.}
\label{tab:rq2-results}
\end{table}

Across all problems, neither augmentation source improves aggregate pass rate. The macro-average pass rate under the baseline is 41.2\%; adding hints yields 40.6\% ($\Delta = -0.6$pp) and adding snippets yields 40.9\% ($\Delta = -0.4$ pp). Snippets outperform hints on 8 of 13 models, hints win on 4, and 1 model ties (Table~\ref{tab:rq2-results}). Neither source improves aggregate accuracy, but snippets have a slight edge in breadth.

\paragraph{Augmentation helps where parametric knowledge is weakest.}
The picture changes substantially on the 219 problems where the zero-shot baseline pass rate is below 25\% (collectively the ``hard'' subset). On these, hints lift the pooled pass rate by $+2.9$pp (12.9\% $\to$ 15.8\%) and snippets by $+3.9$ pp (12.9\% $\to$ 16.9\%). The positive signal is confined to problems where the model's parametric knowledge is sparse enough that external context genuinely fills a gap. In cases where the model already retrieves the algorithm reliably, injecting context provides no additional signal and can introduce noise that displaces correct reasoning.

\paragraph{Snippets vs.\ hints on complex algorithms.}
On the hardest graph-algorithm instances (e.g., Hopcroft--Karp, Stoer-Wagner, Blossom), snippets provide more consistent gains than hints. Within the graph-algorithm domain specifically, snippets lift the pooled pass rate by $+1.4$ pp versus $+0.6$ pp for hints, where a concrete, executable reference is more informative than language-agnostic pseudocode. QwQ-32B, the strongest reasoning model, gains $+5.0$ pp from snippets versus $+4.0$pp from hints, consistent with the view that executable context benefits models already capable of integrating it. This suggests that for low-frequency, structurally complex algorithms whose pseudocode descriptions are themselves specialized, a concrete executable snippet is a more effective retrieval cue than an abstract hint.

\subsection{RQ3: SFT vs. GRPO Fine-Tuning for Closing Language and Representation Gaps}
Prompt augmentation treats retrieval failures as a deployment-time problem, patching them case by case. Fine-tuning with execution-based rewards takes a different approach: adjusting the model's weights so that retrieval becomes more reliable across languages and representations without requiring external context. We apply GRPO~\citep{shao2024deepseekmathpushinglimitsmathematical} to gpt-oss-20B
using execution correctness as the reward signal, and compare the resulting checkpoint against both the base model and a SFT baseline trained on the positive-reward subset of the same training examples.

We fine-tune gpt-oss-20B using LoRA adapters ($r{=}16$, $\alpha{=}32$, dropout $=0.05$) on all seven projection matrices, keeping approximately 0.5\% of parameters trainable. Both GRPO and SFT are trained on the 56-algorithm training split (447 examples spanning all 7 languages and up to 3 input representations per algorithm), with 22 algorithms held out for evaluation. Training examples are organized into representative groups where all examples sharing a canonical algorithm name (across languages and representations) constitute one group. The reward function uses a four-tier ladder: exact stdout match ($r=1.0$), normalized match after whitespace collapsing ($r=0.8$), wrong output with clean execution ($r=0.3$), and any runtime crash or static failure ($r=0.0$--$0.1$). Advantages are computed within each repr group via standard GRPO normalization. The groups with zero reward variance are skipped. The SFT baseline is trained via rejection sampling on the 367 examples where the base model already achieves reward 1.0, using the same LoRA configuration, isolating the effect of policy-gradient credit assignment from the effect of simply training on correct solutions. Full hyperparameters appear in Appendix~\ref{sec:rq3-hyperparams}.

\begin{table}[ht]
\centering
\resizebox{0.8\linewidth}{!}{%
\renewcommand{\arraystretch}{1.2}
\newcommand{\dpos}[1]{\textcolor{teal}{\textbf{$+$#1}}}
\newcommand{\dzer}{\textcolor{gray!70!black}{$0.0$}}
\newcommand{\dneg}[1]{\textcolor{red!70!black}{\textbf{$-$#1}}}
\begin{tabular}{lc@{\hspace{8pt}}ccccccc}
\toprule
& \multicolumn{8}{c}{\textbf{Pass@1 (\%) on held-out test split}} \\
\cmidrule(lr){2-9}
\textbf{Method} & \textbf{Overall} &
\shortstack[c]{\includegraphics[height=1.1em]{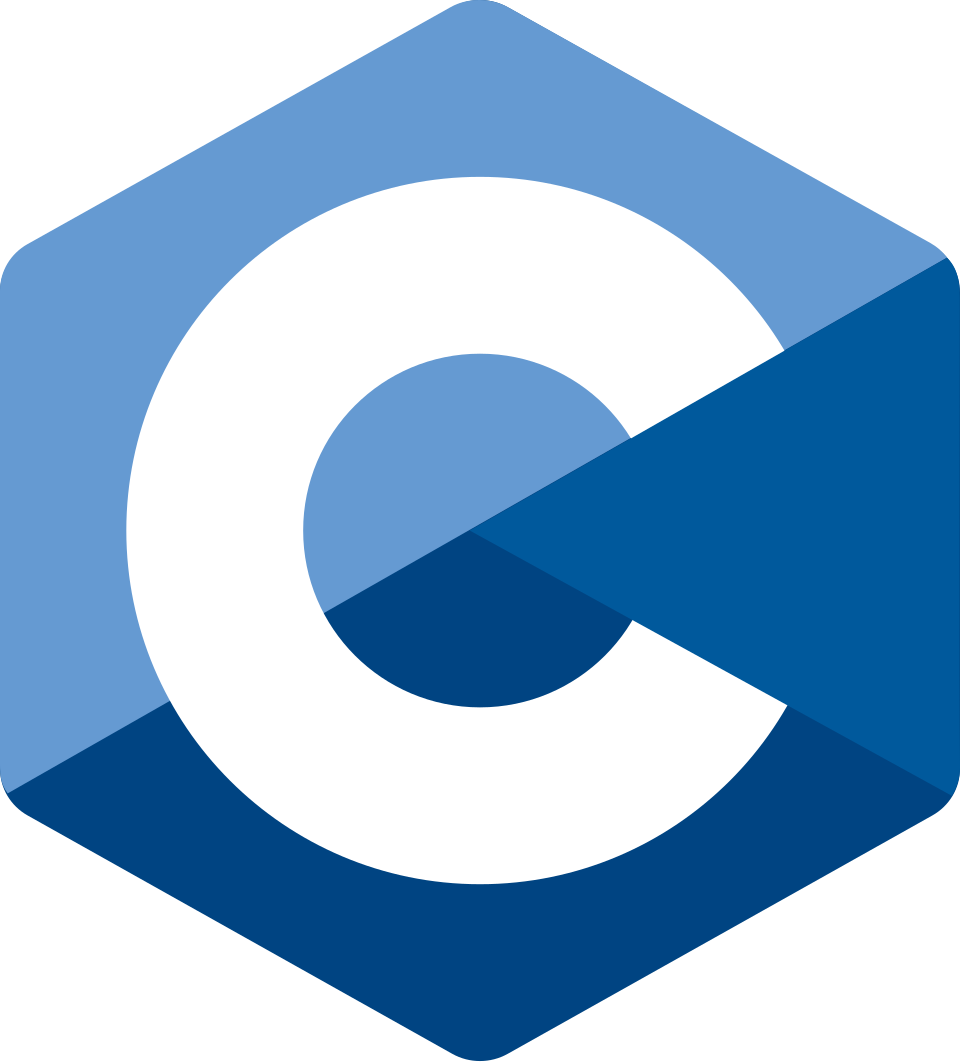}\\[1pt]\textbf{C}} &
\shortstack[c]{\includegraphics[height=1.1em]{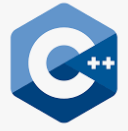}\\[1pt]\textbf{C\texttt{++}}} &
\shortstack[c]{\includegraphics[height=1.1em]{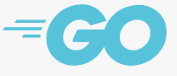}\\[1pt]\textbf{Go}} &
\shortstack[c]{\includegraphics[height=1.1em]{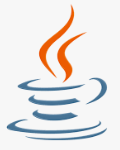}\\[1pt]\textbf{Java}} &
\shortstack[c]{\includegraphics[height=1.1em]{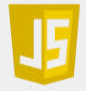}\\[1pt]\textbf{JS}} &
\shortstack[c]{\includegraphics[height=1.1em]{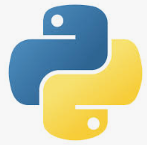}\\[1pt]\textbf{Python}} &
\shortstack[c]{\includegraphics[height=1.1em]{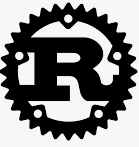}\\[1pt]\textbf{Rust}} \\
\midrule
Base           & 90.1          & 81.0          & 85.7          & \textbf{100.0} & 95.2          & 85.7           & \textbf{96.2} & 85.7          \\
SFT  & \textbf{94.1} & \textbf{85.7} & \textbf{95.2} & \textbf{100.0} & \textbf{100.0} & \textbf{95.2} & \textbf{96.2} & 85.7          \\
GRPO & 92.1          & \textbf{85.7} & 76.2          & \textbf{100.0} & 95.2          & \textbf{100.0} & \textbf{96.2} & \textbf{90.5} \\
\midrule
\textit{$\Delta$ SFT}  & \dpos{3.9}  & \dpos{4.8}  & \dpos{9.5}   & \dzer & \dpos{4.8} & \dpos{9.5}  & \dzer & \dzer       \\
\textit{$\Delta$ GRPO} & \dpos{2.0}  & \dpos{4.8}  & \dneg{9.5}   & \dzer & \dzer      & \dpos{14.3} & \dzer & \dpos{4.8}  \\
\bottomrule
\end{tabular}
}
\caption{\textbf{Pass@1 (\%) on the held-out test split (152 problems, 22 algorithms unseen during
training) for the base model gpt-oss-20b, SFT, and GRPO.} $\Delta$ rows show absolute change vs.\ base in percentage points. SFT leads overall ($+3.9$\,pp) with consistent gains across C, C\texttt{++}, Java, and JavaScript; GRPO achieves the largest JavaScript gain ($+14.3$\,pp) and improves C and Rust,
but regresses on C\texttt{++} ($-9.5$\,pp).}
\label{tab:rq3-results}
\end{table}

GRPO and SFT improve retrieval accuracy in complementary ways (Table~\ref{tab:rq3-results}). SFT achieves broader and more consistent gains, lifting the overall pass rate by $+3.9$\,pp relative to the base model. Because SFT is trained only on already-correct outputs, these gains reflect consolidation: the model reproduces algorithms it has already partially internalised with greater consistency across languages. GRPO's overall gain is smaller ($+2.0$\,pp), but its partial-credit tiers deliver signal that binary supervision cannot. Execution-based rewards add value beyond training on correct solutions alone, but their benefit is concentrated in languages whose failure distribution carries gradable structure. To check for the robustness of splits, we rerun the same SFT/LoRA/rejection-sampling setup on 3 independently-randomized algorithm-level train/test splits. It gives a three-split mean of +0.9pp with a standard deviation (±3.7pp).

\section{Conclusion}

We argue the question worth asking is not whether a model can synthesise a solution but whether it reliably reproduces one it has already internalised.
The most operationally significant failures are wrong-output errors: syntactically valid, runnable code that silently computes the wrong result, invisible without a concrete test case, and present even in the best-performing models. Retrieval failures are partially remediable, but in complementary ways. Prompt augmentation helps only where parametric knowledge is genuinely absent. Execution-based fine-tuning acts at the weight level: SFT achieves broader and more consistent gains across languages ($+3.9$\,pp overall), while GRPO's execution-based reward concentrates signal on specific languages, yielding the largest single-language gain on JavaScript ($+14.3$\,pp) but regressing on C\texttt{++} ($-9.5$\,pp). 


\section*{Limitations}
The results above are subject to three constraints that bound what behavioral evaluation can establish about parametric code retrieval.

A model may pass $t_p$ while silently failing on edge cases such as empty graphs, disconnected components, duplicate keys, or negative-weight cycles, that the fixed input test case does not exercise. EvalPlus~\citep{evalplus} demonstrated that augmenting HumanEval with additional test cases reduces pass rates substantially. A similar augmentation of \datasetname{} would tighten the correctness criterion, but was not feasible at the scale of 599 problems across 7 languages without a verified oracle for each variant.

Our core framing assumes that all 77 algorithms appear in the pretraining corpus of every model we evaluate, and interprets failures as surface-form brittleness rather than genuine knowledge absence. This assumption is untestable: pretraining data composition is rarely disclosed at per-algorithm or per-language granularity. For some low-frequency algorithms in low-resource languages (Go, Rust), what we label a retrieval failure may instead reflect a genuine coverage gap in the pretraining data. Without access to training data, the retrieval and knowledge-absence failure modes cannot be separated.

The paper frames correct model outputs as evidence of parametric retrieval, but external behavior alone cannot fully isolate retrieval from synthesis. A clean separation would require either mechanistic tracing of which training examples causally influence a given generation, or behavioral probes on which a retrieved and a synthesized implementation would make systematically different predictions, neither is available to output-level evaluation.


\clearpage
\appendix

\label{sec:appendix}
\section{Algorithms in \datasetname{}}
\label{sec:algo-list}

Table~\ref{tab:algo-list} lists all 77 algorithms included in \datasetname{}, grouped by domain.

\begin{table*}[ht]
\centering
\resizebox{0.45\linewidth}{!}{%
\begin{tabular}{ll}
\toprule
\textbf{Domain} & \textbf{Algorithm} \\
\midrule
\multirow{3}{*}{Backtracking}
  & N-Queens Problem \\
  & Permutations Generator \\
  & Sudoku Solver \\
\midrule[\lightrulewidth]
\multirow{2}{*}{Compression}
  & Huffman Coding \\
  & Run-Length Encoding \\
\midrule[\lightrulewidth]
\multirow{4}{*}{Computational Geometry}
  & Closest Pair of Points \\
  & Convex Hull \\
  & Point in Polygon \\
  & Polygon Area \\
\midrule[\lightrulewidth]
\multirow{3}{*}{Cryptography}
  & Caesar Cipher \\
  & SHA-256 from Scratch \\
  & Vigenere Cipher \\
\midrule[\lightrulewidth]
\multirow{5}{*}{Data Structures}
  & AVL Tree \\
  & Disjoint Set Union \\
  & Fenwick Tree \\
  & Segment Tree \\
  & Trie \\
\midrule[\lightrulewidth]
\multirow{4}{*}{Dynamic Programming}
  & 0/1 Knapsack \\
  & Edit Distance \\
  & Longest Common Subsequence \\
  & Longest Increasing Subsequence \\
\midrule[\lightrulewidth]
\multirow{24}{*}{Graph Algorithms}
  & Articulation Points \\
  & Bellman-Ford Shortest Path \\
  & Blossom Algorithm --- General Maximum Matching \\
  & Bor\r{u}vka's MST \\
  & Bridge Finding \\
  & Dijkstra's Shortest Path \\
  & Dinic's Max Flow \\
  & Eulerian Path/Circuit \\
  & Floyd-Warshall All-Pairs Shortest Path \\
  & Gabow's SCC \\
  & Hierholzer's Eulerian Circuit \\
  & Hopcroft--Karp \\
  & Johnson's All-Pairs Shortest Paths \\
  & Karger's Min Cut \\
  & Kosaraju's SCC \\
  & Kruskal's MST \\
  & Min-Cost Max-Flow \\
  & Prim's MST \\
  & SCC by Mutual Reachability \\
  & Shortest Path in DAG \\
  & Stoer-Wagner Min Cut \\
  & Tarjan's SCC \\
  & Topological Sort \\
  & Two-SAT \\
\midrule[\lightrulewidth]
\multirow{10}{*}{Image Processing}
  & 2D Convolution with Arbitrary Kernel \\
  & Box Blur (3$\times$3 Averaging Filter) \\
  & Gaussian Blur 3$\times$3 \\
  & Histogram Equalization \\
  & Median Filter 3$\times$3 \\
  & Otsu Thresholding \\
  & Prewitt Edge Detection \\
  & Rotate Image 90$^\circ$ Clockwise \\
  & Rotate Image 180$^\circ$ \\
  & Sobel Edge Detection \\
\midrule[\lightrulewidth]
\multirow{3}{*}{Number Theory}
  & Extended GCD \\
  & Modular Exponentiation \\
  & Sieve of Eratosthenes \\
\midrule[\lightrulewidth]
\multirow{4}{*}{Numerical Methods}
  & Fast Fourier Transform \\
  & Gaussian Elimination \\
  & Newton-Raphson Method \\
  & Simpson's Rule \\
\midrule[\lightrulewidth]
\multirow{6}{*}{Other}
  & Banker's Algorithm for Deadlock Avoidance \\
  & FCFS CPU Scheduling \\
  & FIFO Page Replacement \\
  & LRU Page Replacement \\
  & Non-Preemptive SJF CPU Scheduling \\
  & Round Robin CPU Scheduling \\
\midrule[\lightrulewidth]
\multirow{2}{*}{Parsing}
  & Postfix Expression Evaluator \\
  & Shunting-Yard Algorithm \\
\midrule[\lightrulewidth]
\multirow{4}{*}{Sorting Algorithms}
  & Counting Sort \\
  & Heap Sort \\
  & Merge Sort \\
  & Radix Sort \\
\midrule[\lightrulewidth]
\multirow{3}{*}{String Algorithms}
  & KMP Pattern Search \\
  & Rabin-Karp Pattern Search \\
  & Z-Algorithm Pattern Search \\
\bottomrule
\end{tabular}%
}
\caption{All 77 algorithms in \datasetname{}, grouped by domain (14 domains total).}
\label{tab:algo-list}
\end{table*}

\section{Models}
\label{sec:models}

Table~\ref{tab:models} lists all 15 models evaluated in this work, along with their HuggingFace model IDs (or API identifiers for closed models), parameter counts, organizations, and licenses.

\begin{table*}[htbp]
\centering
\resizebox{\linewidth}{!}{%
\begin{tabular}{llrll}
\toprule
\textbf{Model} & \textbf{Model ID} & \textbf{Params} & \textbf{Organization} & \textbf{License} \\
\midrule
CodeGemma-7B           & \texttt{google/codegemma-7b-it}                       &  7B & Google     & Gemma \\
CodeLlama-34B          & \texttt{codellama/CodeLlama-34b-Instruct-hf}           & 34B & Meta       & Llama 2 \\
Codestral-22B          & \texttt{mistralai/Codestral-22B-v0.1}                  & 22B & Mistral AI & MNPL \\
DeepSeek-Coder-33B     & \texttt{deepseek-ai/deepseek-coder-33b-instruct}       & 33B & DeepSeek   & DeepSeek \\
DeepSeek-R1-Llama-8B   & \texttt{deepseek-ai/DeepSeek-R1-Distill-Llama-8B}     &  8B & DeepSeek   & MIT \\
DeepSeek-R1-Qwen-14B   & \texttt{deepseek-ai/DeepSeek-R1-Distill-Qwen-14B}     & 14B & DeepSeek   & MIT \\
GPT-4.1-mini           & \texttt{gpt-4.1-mini-2025-04-14} (API)                 & --- & OpenAI     & Proprietary \\
GPT-4o-mini            & \texttt{gpt-4o-mini-2024-07-18} (API)                  & --- & OpenAI     & Proprietary \\
gpt-oss-20B            & \texttt{openai/gpt-oss-20b}                            & 20B & OpenAI     & MIT \\
Llama-3.1-8B           & \texttt{meta-llama/Llama-3.1-8B-Instruct}             &  8B & Meta       & Llama 3 \\
Phi-4                  & \texttt{microsoft/phi-4}                               & 14B & Microsoft  & MIT \\
QwQ-32B                & \texttt{Qwen/QwQ-32B}                                  & 32B & Alibaba    & Apache 2.0 \\
Qwen2.5-7B             & \texttt{Qwen/Qwen2.5-7B-Instruct}                     &  7B & Alibaba    & Apache 2.0 \\
StarCoder2-15B         & \texttt{bigcode/starcoder2-15b-instruct-v0.1}          & 15B & BigCode    & OpenRAIL-M \\
Yi-Coder-9B            & \texttt{01-ai/Yi-Coder-9B-Chat}                        &  9B & 01.AI      & Apache 2.0 \\
\bottomrule
\end{tabular}%
}
\caption{All 15 models evaluated in \datasetname{}. Open-weight models were run locally via vLLM. Closed models were accessed through their respective APIs. Parameter counts for closed API models are not publicly disclosed.}
\label{tab:models}
\end{table*}

\section{Zero-Shot Inference Hyperparameters}
\label{sec:hyperparams}

Table~\ref{tab:hyperparams} summarises the generation parameters used.
Open-weight models were served locally via vLLM and run with greedy decoding (temperature~$= 0$, top-$p = 1.0$).
Closed API models (GPT-4o-mini, GPT-4.1-mini) were queried through the OpenAI API with temperature $= 0.7$ and a smaller token budget, as the API enforces its own safety caps.
Thinking models (QwQ-32B, DeepSeek-R1-Distill-Llama-8B, DeepSeek-R1-Distill-Qwen-14B) share the vLLM settings but require a large output budget to avoid truncating the internal reasoning trace before code is emitted; the 32{,}768-token limit serves as both the thinking budget and the code output window.
A per-problem execution timeout of 30 seconds was applied uniformly across all models and languages.

\begin{table}[htbp]
\centering
\resizebox{1.\linewidth}{!}{%
\begin{tabular}{lcc}
\toprule
\textbf{Parameter} & \textbf{Open-weight (vLLM)} & \textbf{Closed API} \\
\midrule
Temperature        & 0 (greedy)   & 0.7 \\
Max output tokens  & 32{,}768     & 8{,}192 \\
Top-$p$            & 1.0          & --- \\
Decoding           & Greedy       & Sampling \\
Thinking budget    & 32{,}768 (thinking models) & N/A \\
Execution timeout  & 30 s         & 30 s \\
Prompting strategy & Zero-shot    & Zero-shot \\
\bottomrule
\end{tabular}
}
\caption{Generation hyperparameters for the zero-shot evaluation. Open-weight models include all 13 locally-served models. Closed API covers GPT-4o-mini and GPT-4.1-mini. The thinking-model token floor applies to QwQ-32B, DeepSeek-R1-Distill-Llama-8B, and DeepSeek-R1-Distill-Qwen-14B.}
\label{tab:hyperparams}
\end{table}

\section{Prompt-Augmentation Setup}
\label{sec:rq2-hyperparams}

It tests whether retrieval failures can be remediated by injecting external context into the prompt.
We evaluate two conditions using \textbf{+hint} and \textbf{+snippet} across the same 13 open-weight vLLM models used in the zero-shot evaluation (no closed-API models).
Generation hyperparameters are identical to zero-shot: temperature~$= 0$, max output tokens~$= 32{,}768$, top-$p = 1.0$, execution timeout $= 30$\,s.

\paragraph{+Hint condition.}
Structured algorithmic hints were generated offline using \texttt{gpt-4o-mini-2024-07-18} at a temperature of $0$.
Each hint covers three fields: (1)~language-agnostic pseudocode ($\leq$20 lines), (2)~key data structures, and (3)~Big-O time/space complexity.
Because the hints are language-agnostic, the same hint is prepended to all seven language variants of a given algorithm.
Hints were generated for all 77 unique algorithms.

\paragraph{+Snippet condition.}
One reference implementation per (algorithm, language) pair was retrieved from GitHub using the GitHub Search API (10 results per query, sorted by index date). Results from RosettaCode and GeeksforGeeks were excluded to avoid overlap with the benchmark's source material.

Each retrieved file was truncated to at most 80 lines, centred on the most keyword-relevant region. A 2.2\,s inter-request delay was used to respect the GitHub API's rate limit. Coverage-wise, 431 of 599 problems received an injected snippet. The remaining 168 (no language-matched result found) fell back to the unaugmented baseline prompt.

\paragraph{Prompt injection.}
In both conditions, the system message remains unchanged relative to the zero-shot baseline. The injected block is inserted immediately before the stdlib-constraint footer, so it is not buried after the final instruction.

The snippet block uses the prefix: \textit{``For reference, here is an implementation of this algorithm in \{lang\} retrieved from a different source. You may use it as a reference but must write your own solution.''} followed by a fenced code block.

The hint block uses the header \textit{``Algorithmic hint:''} followed by the pseudocode, data-structure list, and complexity note.
Table~\ref{tab:rq2-hyperparams} summarises the setup.

\begin{table}[htbp]
\centering
\resizebox{\linewidth}{!}{%
\begin{tabular}{lll}
\toprule
\textbf{Parameter} & \textbf{+Hint} & \textbf{+Snippet} \\
\midrule
Models evaluated       & 13 vLLM (open-weight)           & 13 vLLM (open-weight) \\
Temperature            & 0 (greedy)                      & 0 (greedy) \\
Max output tokens      & 32{,}768                        & 32{,}768 \\
Top-$p$                & 1.0                             & 1.0 \\
Execution timeout      & 30 s                            & 30 s \\
Context source         & GPT-4o-mini (temp.\,0)          & GitHub Search API \\
Context granularity    & Per algorithm (lang-agnostic)   & Per (algorithm, language) \\
Max context length     & $\leq$20 pseudocode lines       & $\leq$80 code lines \\
Problem coverage       & 599/599                         & 599/599 \\
Excluded sources       & N/A                             & RosettaCode, GeeksforGeeks \\
\bottomrule
\end{tabular}
}
\caption{Prompt-augmentation parameters. Generation settings are identical to the zero-shot baseline. Only the prompt contents differ between conditions.}
\label{tab:rq2-hyperparams}
\end{table}

\section{Fine-Tuning Hyperparameters}
\label{sec:rq3-hyperparams}

Both GRPO and SFT fine-tune gpt-oss-20B using LoRA adapters applied to all seven projection matrices
($q$, $k$, $v$, $o$, gate, up, down), keeping approximately 0.5\% of parameters trainable.
Table~\ref{tab:rq3-hyperparams} lists the shared and condition-specific hyperparameters.

\begin{table}[htbp]
\centering
\resizebox{\linewidth}{!}{%
\begin{tabular}{lll}
\toprule
\textbf{Parameter} & \textbf{GRPO} & \textbf{SFT} \\
\midrule
Base model              & gpt-oss-20B              & gpt-oss-20B \\
LoRA rank $r$           & 16                        & 16 \\
LoRA $\alpha$           & 32                        & 32 \\
LoRA dropout            & 0.05                      & 0.05 \\
Trainable params        & $\approx$0.5\%            & $\approx$0.5\% \\
Training algorithms     & 56 (447 examples)         & 56 (367 positive examples) \\
Training examples       & All 7 langs, $\leq$3 repr & Rejection-sampled ($r{=}1.0$) \\
Held-out test split     & 22 algorithms, 152 problems & 22 algorithms, 152 problems \\
Grouping                & Repr groups (algo $\times$ lang $\times$ repr) & N/A \\
Reward tiers            & 1.0 / 0.8 / 0.3 / 0.0--0.1 & N/A \\
Optimiser               & AdamW                     & AdamW \\
Epochs                  & 3                         & 3 \\
Execution timeout       & 30 s                      & 30 s \\
\bottomrule
\end{tabular}
}
\caption{Fine-tuning hyperparameters for GRPO and SFT. Both conditions use the same LoRA configuration and train/test split. SFT trains only on examples where the base model already achieves $r{=}1.0$ (rejection sampling), isolating the contribution of policy-gradient credit assignment from the effect of training on correct solutions alone.}
\label{tab:rq3-hyperparams}
\end{table}

\section{Prompt Template}
\label{sec:prompt-template}

All models are prompted with a fixed system message followed by a per-problem user message. Figure~\ref{fig:prompt-example} shows a representative example for the Hopcroft--Karp algorithm in Python (edge-list representation).

\begin{figure*}[ht]
\centering
\begin{tcolorbox}[
  colback=gray!5,
  colframe=gray!40,
  title={\small\textbf{System message}},
  fonttitle=\small,
  left=4pt, right=4pt, top=4pt, bottom=4pt
]
\small\ttfamily
You are an expert software engineer tasked with implementing algorithms correctly and efficiently.

Output ONLY a single fenced code block with the complete, self-contained program. No explanations, no prose, no install commands --- just the code block.

Requirements:
\begin{itemize}[nosep,leftmargin=1.2em]
  \item The program must run/compile as-is without any modifications.
  \item Use only the language's standard library.
  \item Do not add any extra output beyond what the problem specifies.
  \item The code must be in a single file.
\end{itemize}
\textit{[language-specific notes omitted for brevity]}
\end{tcolorbox}

\vspace{4pt}

\begin{tcolorbox}[
  colback=blue!3,
  colframe=blue!30,
  title={\small\textbf{User message} (Hopcroft--Karp, Python, edge-list)},
  fonttitle=\small,
  left=4pt, right=4pt, top=4pt, bottom=4pt
]
\small\ttfamily
Implement the Hopcroft--Karp algorithm for maximum matching in a bipartite graph in \textbf{Python}.

Requirements:
\begin{itemize}[nosep,leftmargin=1.2em]
  \item Provide a class \texttt{BipGraph(m, n)} with:
  \begin{itemize}[nosep,leftmargin=1.2em]
    \item \texttt{addEdge(u, v)} using 1-based indices (u in U\,=\,1..m, v in V\,=\,1..n)
    \item \texttt{hopcroftKarp()} returning the integer size of the maximum matching.
  \end{itemize}
  \item The code must be runnable as a script.
  \item Include the following exact test case under \texttt{if \_\_name\_\_ == "\_\_main\_\_":} (do not change it):
\end{itemize}

\begin{verbatim}
if __name__ == "__main__":
    g = BipGraph(4, 4)
    g.addEdge(1, 2)
    g.addEdge(1, 3)
    g.addEdge(2, 1)
    g.addEdge(3, 2)
    g.addEdge(4, 2)
    g.addEdge(4, 4)
    print("Size of maximum matching is %d"
          % g.hopcroftKarp())
\end{verbatim}

Runtime constraints (strict):
\begin{itemize}[nosep,leftmargin=1.2em]
  \item Use only the language standard library.
  \item Do NOT use third-party packages or package managers.
  \item Output only executable code.
\end{itemize}
\end{tcolorbox}
\caption{Prompt structure used for all evaluations. The system message is fixed across all problems and languages. The user message is generated per problem; it names the algorithm and target language, specifies the required interface, and embeds an executable test case whose output is compared against the ground-truth reference output.}
\label{fig:prompt-example}
\end{figure*}
\section{Prompt Augmentation Examples}
\label{sec:augmentation-examples}

Figures~\ref{fig:prompt-hint} and~\ref{fig:prompt-snippet} show the complete user
messages fed to models under each RQ2 augmentation condition for the
Hopcroft--Karp (Python) example. The system message is identical to
Figure~\ref{fig:prompt-example} in all conditions. The augmentation block
(shaded region) is inserted immediately before the stdlib-constraint footer so
the injected context is visible before the final instruction.

\begin{figure*}[ht]
\centering
\begin{adjustbox}{width=0.7\linewidth}
\begin{tcolorbox}[
  colback=blue!3,
  colframe=blue!30,
  title={\small\textbf{User message --- \texttt{+Hint} condition
         (Hopcroft--Karp, Python)}},
  fonttitle=\small,
  left=4pt, right=4pt, top=4pt, bottom=4pt
]
\small\ttfamily
Implement the Hopcroft--Karp algorithm for maximum matching in a
bipartite graph in \textbf{Python}.

Requirements:
\begin{itemize}[nosep,leftmargin=1.2em]
  \item Provide a class \texttt{BipGraph(m, n)} with:
  \begin{itemize}[nosep,leftmargin=1.2em]
    \item \texttt{addEdge(u, v)} using \textbf{1-based} indices
          (u in U\,=\,1..m, v in V\,=\,1..n)
    \item \texttt{hopcroftKarp()} returning the integer size of the
          maximum matching.
  \end{itemize}
  \item The code must be runnable as a script.
  \item Include the following exact test case \textbf{under}
        \texttt{if \_\_name\_\_ == "\_\_main\_\_":} (do not change it):
\end{itemize}

\begin{verbatim}
if __name__ == "__main__":
    g = BipGraph(4, 4)
    g.addEdge(1, 2)
    g.addEdge(1, 3)
    g.addEdge(2, 1)
    g.addEdge(3, 2)
    g.addEdge(4, 2)
    g.addEdge(4, 4)
    print("Size of maximum matching is %d" % g.hopcroftKarp())
\end{verbatim}

\begin{tcolorbox}[
  colback=gray!5,
  colframe=gray!40,
  title={\small\textit{Injected hint block}},
  fonttitle=\small\itshape,
  left=3pt, right=3pt, top=3pt, bottom=3pt,
  boxrule=0.4pt
]
\small
\begin{verbatim}
Algorithmic hint:
Pseudocode:
1. Initialize matching M to empty
2. While there exists an augmenting path P in the graph:
   2.1. Augment the matching M along path P
3. Return matching M
4. Function to find augmenting path P:
   4.1. Initialize queue Q and visited set V
   4.2. For each vertex u in the left set:
       4.2.1. If u is unmatched, add u to Q and V
   4.3. While Q is not empty:
       4.3.1. Dequeue vertex u from Q
       4.3.2. For each neighbor v of u in the right set:
           4.3.2.1. If v is unmatched or matched vertex of v
                    can find an augmenting path:
               4.3.2.1.1. Add v to Q and V
               4.3.2.1.2. Record the path from u to v
   4.4. If an augmenting path is found, return it
   4.5. Otherwise, return null
Key data structures: queue, visited set, matching set
Complexity: O(E*sqrt(V)) time, O(V) space
\end{verbatim}
\end{tcolorbox}

\vspace{4pt}
\small\ttfamily
Runtime constraints (strict):
\begin{itemize}[nosep,leftmargin=1.2em]
  \item Use only the language standard library.
  \item Do NOT use third-party packages/crates/modules.
  \item Do NOT use package managers (pip/npm/cargo/maven/gradle/etc.).
  \item If a data structure/algorithm is missing, implement it directly
        in code.
  \item Output only executable code.
\end{itemize}
\end{tcolorbox}
\end{adjustbox}
\caption{Complete user message for Hopcroft--Karp (Python) under the
\textbf{+Hint} condition. The green-shaded block is the only addition
relative to the zero-shot baseline; the same language-agnostic hint is
used for all seven language variants of the algorithm. Generated offline
by \texttt{gpt-4o-mini-2024-07-18} at temperature~$= 0$.}
\label{fig:prompt-hint}
\end{figure*}

\begin{figure*}[ht]
\centering
\begin{adjustbox}{width=0.6\linewidth}
\begin{tcolorbox}[
  colback=blue!3,
  colframe=blue!30,
  title={\small\textbf{User message --- \texttt{+Snippet} condition
         (Hopcroft--Karp, Python)}},
  fonttitle=\small,
  left=4pt, right=4pt, top=4pt, bottom=4pt
]
\small\ttfamily
Implement the Hopcroft--Karp algorithm for maximum matching in a
bipartite graph in \textbf{Python}.

Requirements:
\begin{itemize}[nosep,leftmargin=1.2em]
  \item Provide a class \texttt{BipGraph(m, n)} with:
  \begin{itemize}[nosep,leftmargin=1.2em]
    \item \texttt{addEdge(u, v)} using \textbf{1-based} indices
          (u in U\,=\,1..m, v in V\,=\,1..n)
    \item \texttt{hopcroftKarp()} returning the integer size of the
          maximum matching.
  \end{itemize}
  \item The code must be runnable as a script.
  \item Include the following exact test case \textbf{under}
        \texttt{if \_\_name\_\_ == "\_\_main\_\_":} (do not change it):
\end{itemize}

\begin{verbatim}
if __name__ == "__main__":
    g = BipGraph(4, 4)
    g.addEdge(1, 2)
    g.addEdge(1, 3)
    g.addEdge(2, 1)
    g.addEdge(3, 2)
    g.addEdge(4, 2)
    g.addEdge(4, 4)
    print("Size of maximum matching is %d" % g.hopcroftKarp())
\end{verbatim}

\begin{tcolorbox}[
  colback=gray!5,
  colframe=gray!40,
  title={\small\textit{Injected snippet block}},
  fonttitle=\small\itshape,
  left=3pt, right=3pt, top=3pt, bottom=3pt,
  boxrule=0.4pt
]
\small
\begin{verbatim}
For reference, here is an implementation of this algorithm in python
retrieved from a different source. You may use it as a reference but
must write your own solution:
```python
"""BipartiteMatching.py

Hopcroft-Karp bipartite maximum-cardinality matching and maximum
independent set for bipartite graphs.

D. Eppstein, April 2002.
"""

from .strong_connectivity import StronglyConnectedComponents


def matching(graph):
    """Find maximum cardinality matching of a bipartite graph (U,V,E).
    The input format is a dictionary mapping members of U to lists
    of their neighbors in V.
    """
    # initialize greedy matching (redundant, but faster than full search)
    matching = {}
    for u in graph:
        for v in graph[u]:
            if v not in matching:
                matching[v] = u
                break

    while True:
        # structure residual graph into layers
        preds = {}
        unmatched = []
        pred = {u: unmatched for u in graph}
        for v in matching:
            del pred[matching[v]]
        layer = list(pred)

        # repeatedly extend layering structure by another pair of layers
        while layer and not unmatched:
            newLayer = {}
            for u in layer:
                for v in graph[u]:
                    if v not in preds:
                        newLayer.setdefault(v, []).append(u)
            layer = []
            for v in newLayer:
                preds[v] = newLayer[v]
                if v in matching:
                    layer.append(matching[v])
                    pred[matching[v]] = v
                else:
                    unmatched.append(v)

        # did we finish layering without finding any alternating paths?
        if not unmatched:
            unlayered = {}
            for u in graph:
                for v in graph[u]:
                    if v not in preds:
                        unlayered[v] = None
            return (matching, list(pred), list(unlayered))

        # recursively search backward through layers to find alternating paths
        def recurse(v):
            if v in preds:
                L = preds[v]
                del preds[v]
                for u in L:
                    if u in pred:
                        pu = pred[u]
                        del pred[u]
                        if pu is unmatched or recurse(pu):
```
\end{verbatim}
\end{tcolorbox}

\vspace{4pt}
\small\ttfamily
Runtime constraints (strict):
\begin{itemize}[nosep,leftmargin=1.2em]
  \item Use only the language standard library.
  \item Do NOT use third-party packages/crates/modules.
  \item Do NOT use package managers (pip/npm/cargo/maven/gradle/etc.).
  \item If a data structure/algorithm is missing, implement it directly
        in code.
  \item Output only executable code.
\end{itemize}
\end{tcolorbox}
\end{adjustbox}
\caption{Complete user message for Hopcroft--Karp (Python) under the
\textbf{+Snippet} condition. The orange-shaded block is the only addition
relative to the zero-shot baseline. The snippet (D.\ Eppstein, 2002) was
retrieved from GitHub via the GitHub Search API; RosettaCode and
GeeksforGeeks results were excluded to avoid overlap with the benchmark's
source material. Files are truncated to at most 80 lines centred on the
most keyword-relevant region, hence the snippet ends mid-function. Each
language variant receives its own language-matched snippet.}
\label{fig:prompt-snippet}
\end{figure*}

\section{Error Analysis}
\label{sec:error-analysis}

\begin{figure*}[t]
    \centering
    \includegraphics[width=0.75\linewidth]{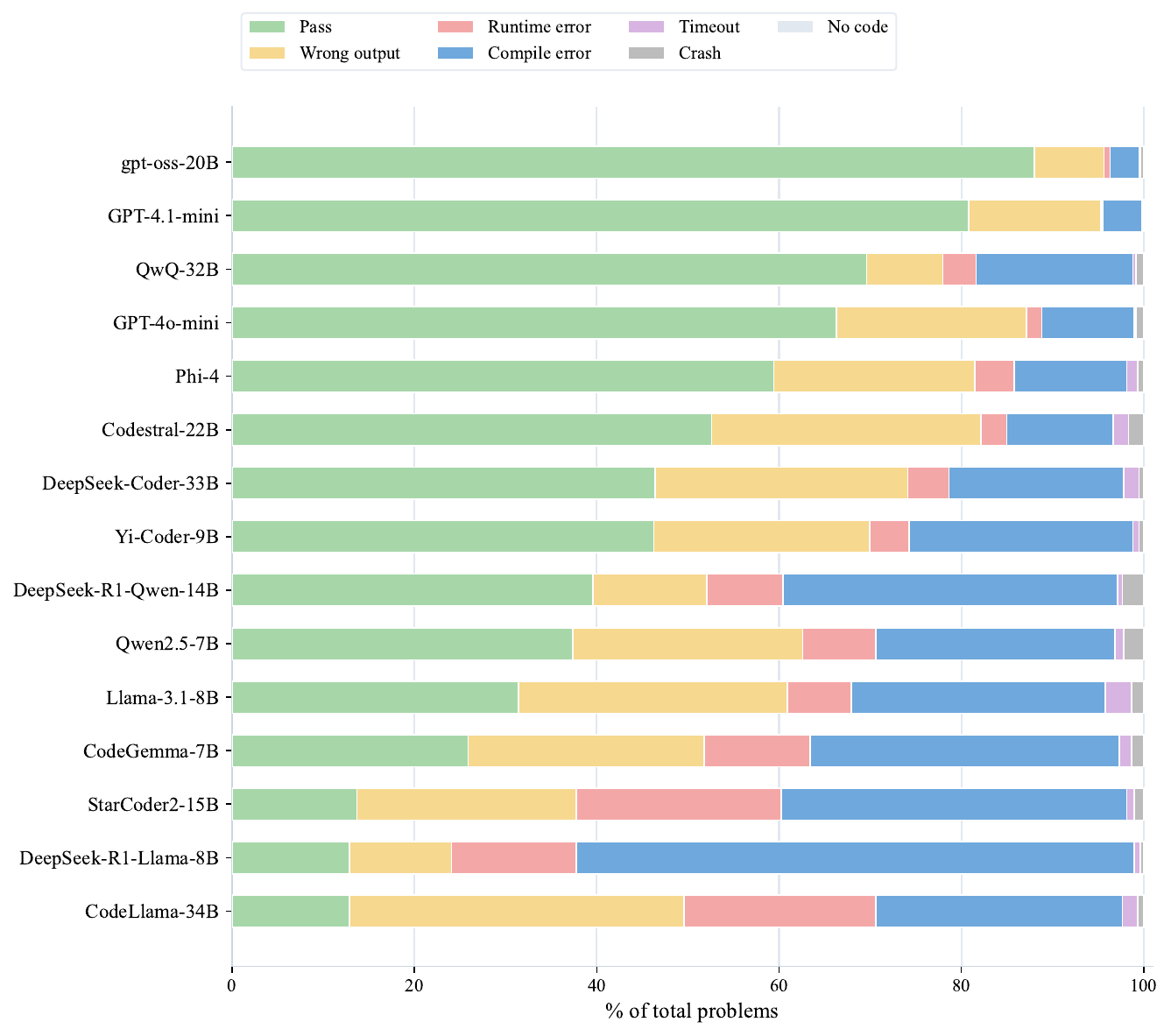}
    \caption{Outcome breakdown per model as a fraction of all 599 problems, sorted by pass rate.
    Stronger models fail almost exclusively on wrong output, while weaker models are dominated by compile errors,
    reflecting a gradient from syntactic to semantic retrieval failure.}
    \label{fig:error-analysis}
\end{figure*}

\begin{figure*}[t]
    \centering
    \includegraphics[width=0.75\linewidth]{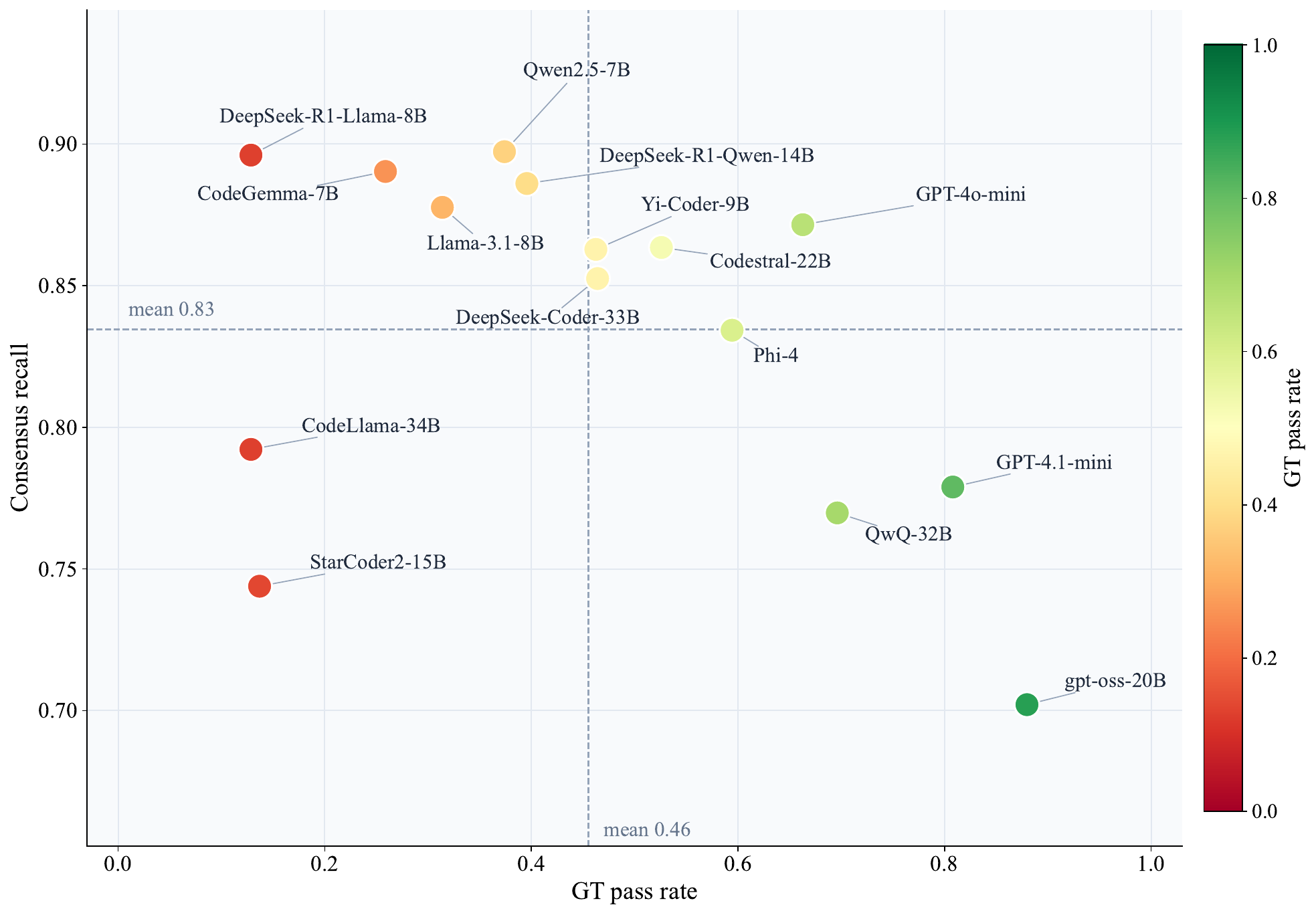}
    \caption{Pass rate vs.\ consensus recall. GT pass rate (x-axis) is the fraction of problems where a model's output matches the ground-truth expected value. Consensus recall (y-axis) is the fraction of a model's correct answers that also match the plurality output across all models.}
    \label{fig:consensus-recall}
\end{figure*}

Figure~\ref{fig:error-analysis} breaks down evaluation outcomes per model across all 599 problems.
Stronger models fail almost exclusively on wrong-output errors, while weaker models are dominated by compile errors, reflecting a gradient from syntactic to semantic retrieval failure.
Figure~\ref{fig:consensus-recall} plots pass rate against consensus recall, showing that models producing correct outputs do not uniformly converge on the same answer — consistent with independent retrieval rather than recitation of a shared token sequence.

\section{Solution Diversity by Model and Language}
\label{sec:diversity-model-lang}
\begin{figure*}[t]
    \centering
    \includegraphics[width=0.75\linewidth]{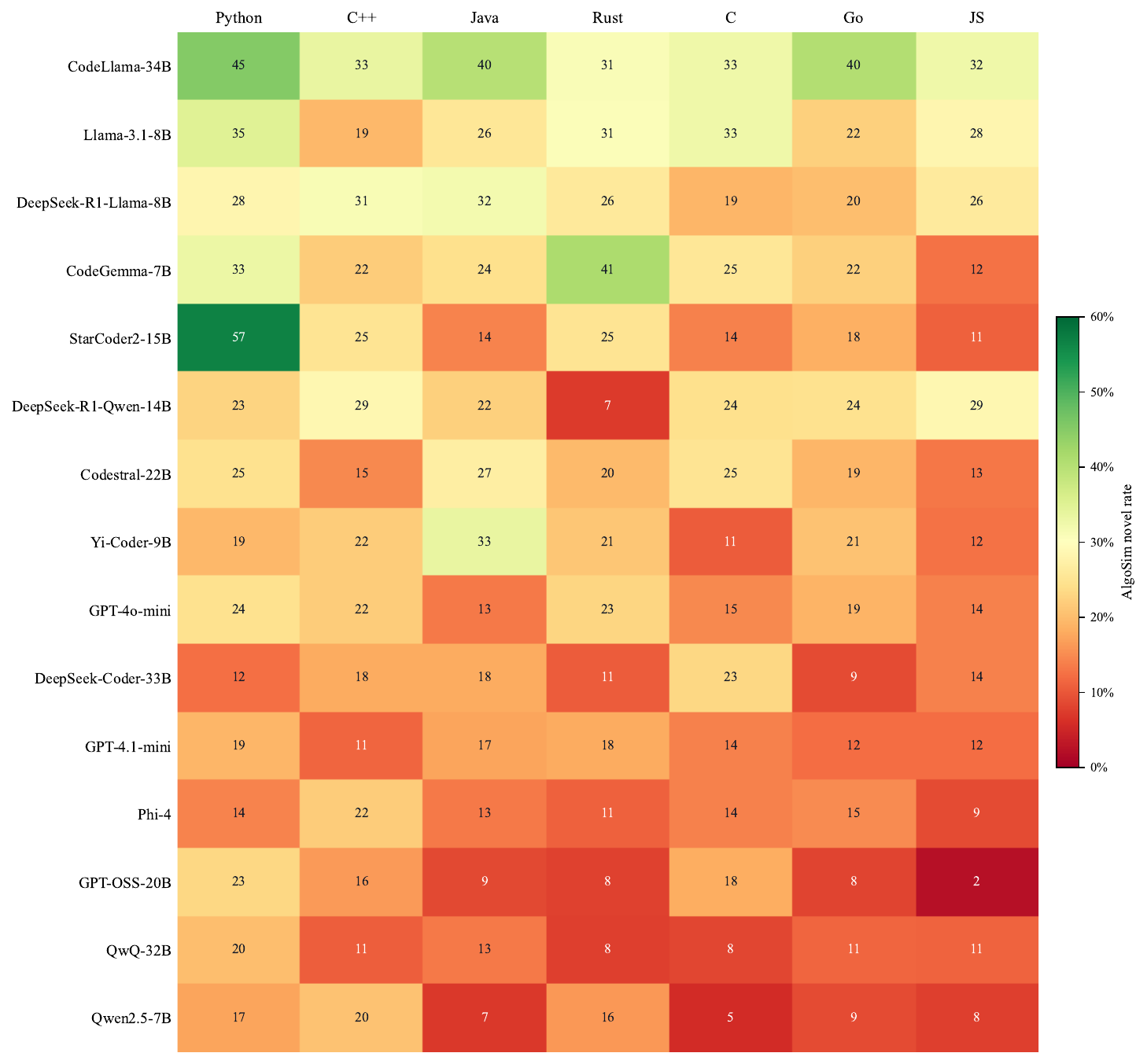}
    \caption{AlgoSim novel rate by model and programming language.}
    \label{fig:diversity-model-lang}
\end{figure*}

Figure~\ref{fig:diversity-model-lang} shows the AlgoSim novel rate per model and language, computed over 2{,}215 cross-model pairwise comparisons across 40 algorithms, 7 languages, and 15 models ($\binom{15}{2}{=}105$ pairs per problem), judged by GPT-4.1-mini. 20.5\% of pairs were judged novel overall.

\section{Per-Language Algorithm Difficulty Heatmaps}
\label{sec:heatmaps}

\begin{figure*}
  \includegraphics[width=0.85\linewidth]{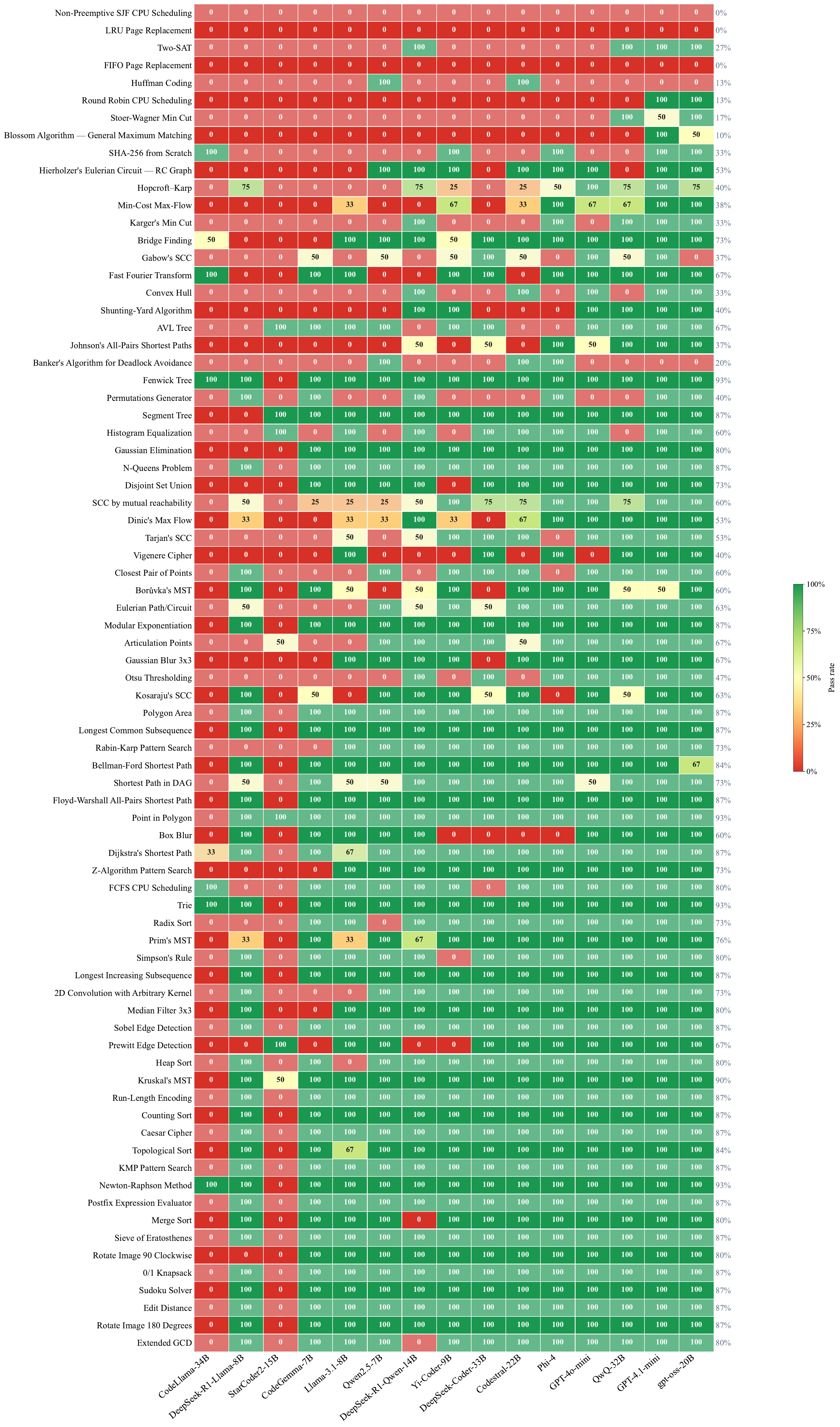}
  \caption{Algorithm difficulty heatmap for Python.}
  \label{fig:heatmap-python}
\end{figure*}

\begin{figure*}
  \includegraphics[width=0.85\linewidth]{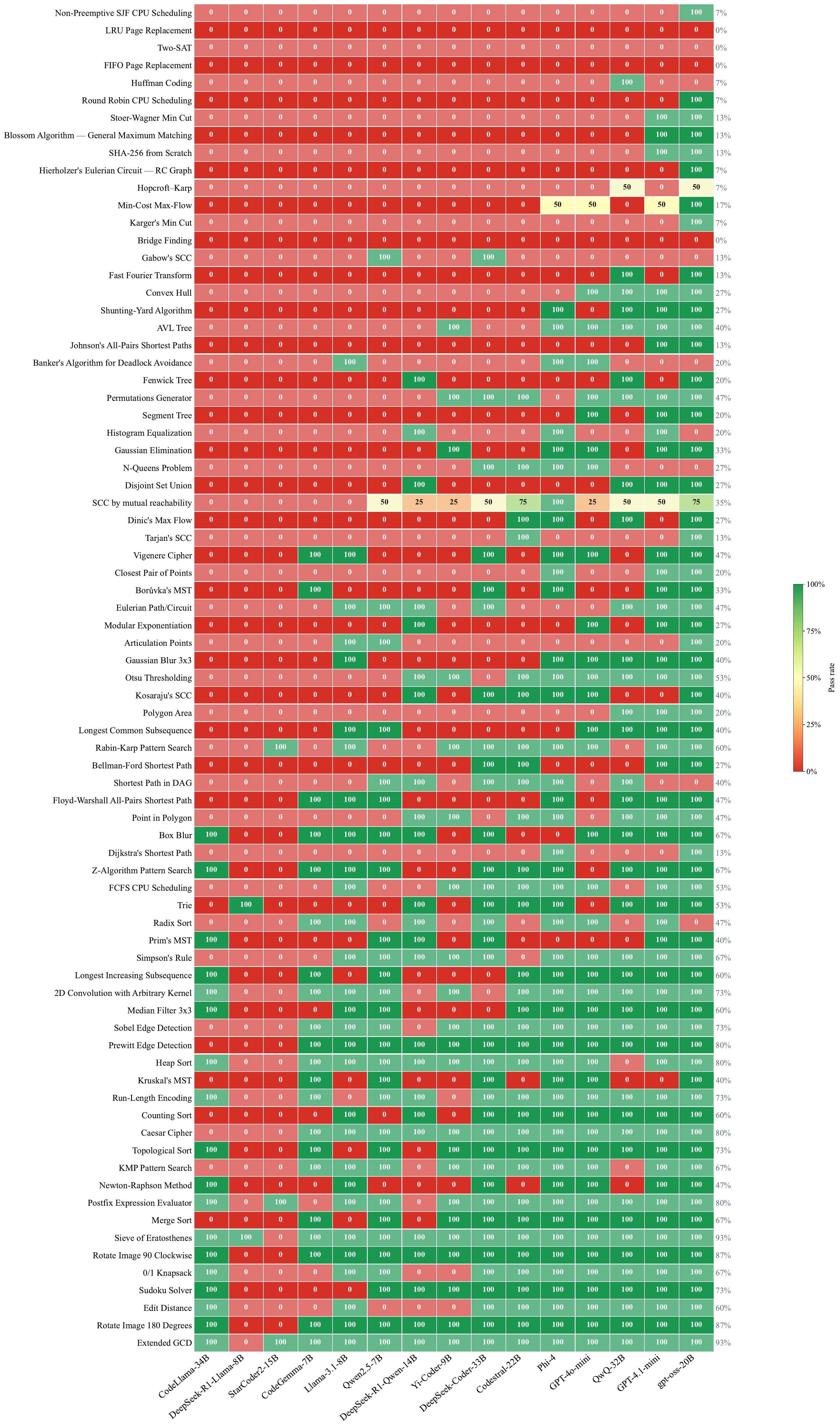}
  \caption{Algorithm difficulty heatmap for C++.}
  \label{fig:heatmap-cpp}
\end{figure*}

\begin{figure*}
  \includegraphics[width=0.85\linewidth]{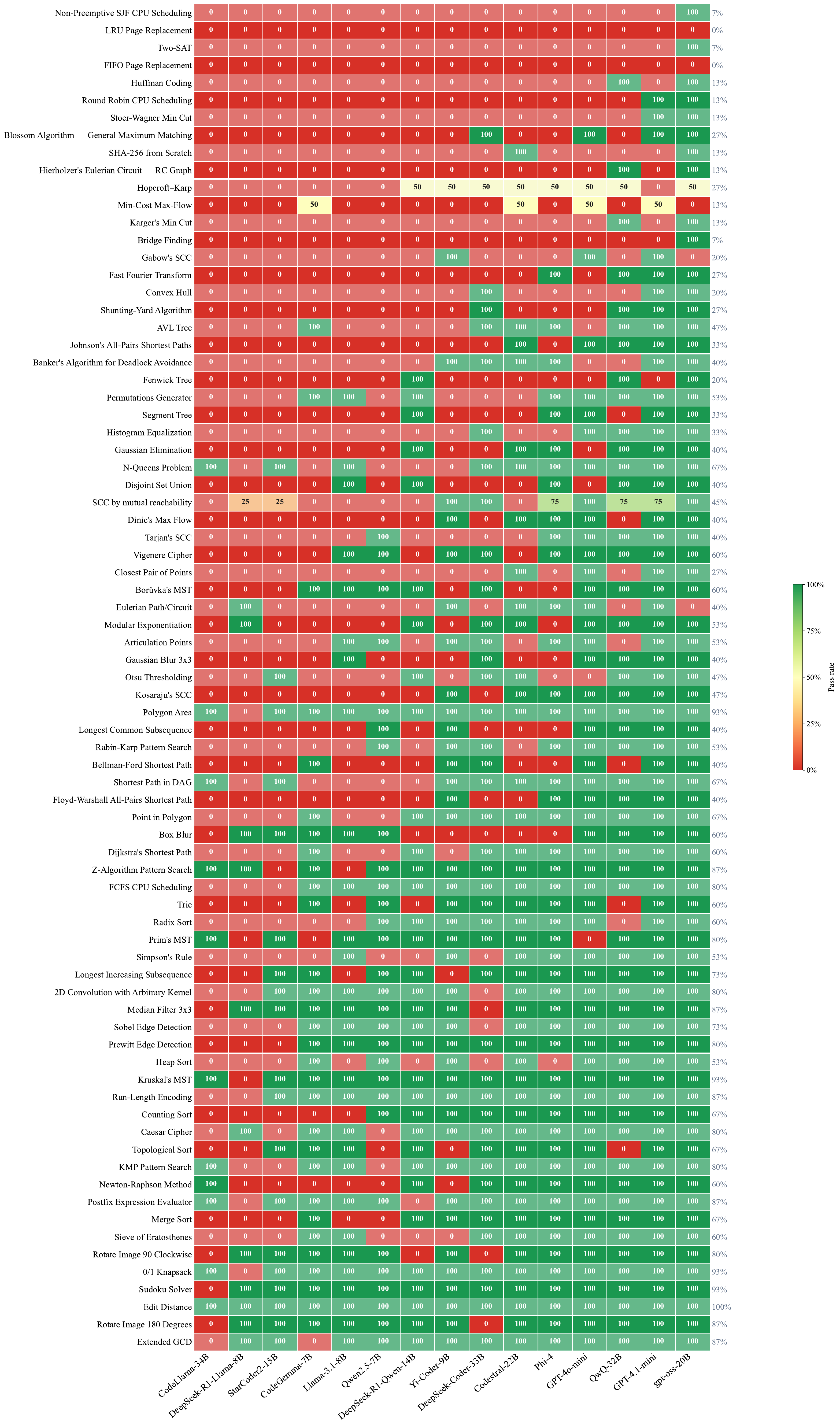}
  \caption{Algorithm difficulty heatmap for Java.}
  \label{fig:heatmap-java}
\end{figure*}

\begin{figure*}
  \includegraphics[width=0.85\linewidth]{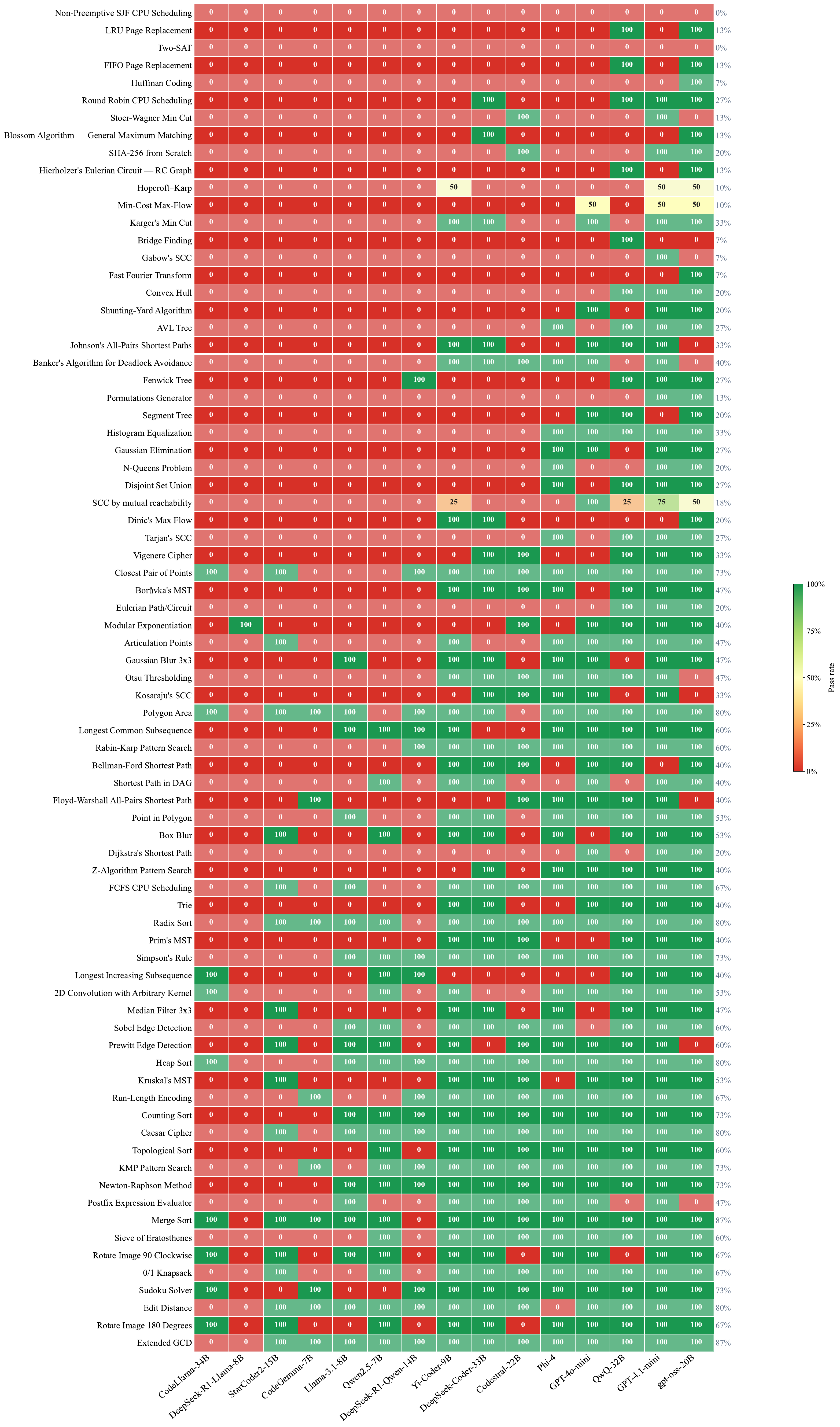}
  \caption{Algorithm difficulty heatmap for C.}
  \label{fig:heatmap-c}
\end{figure*}

\begin{figure*}
  \includegraphics[width=0.85\linewidth]{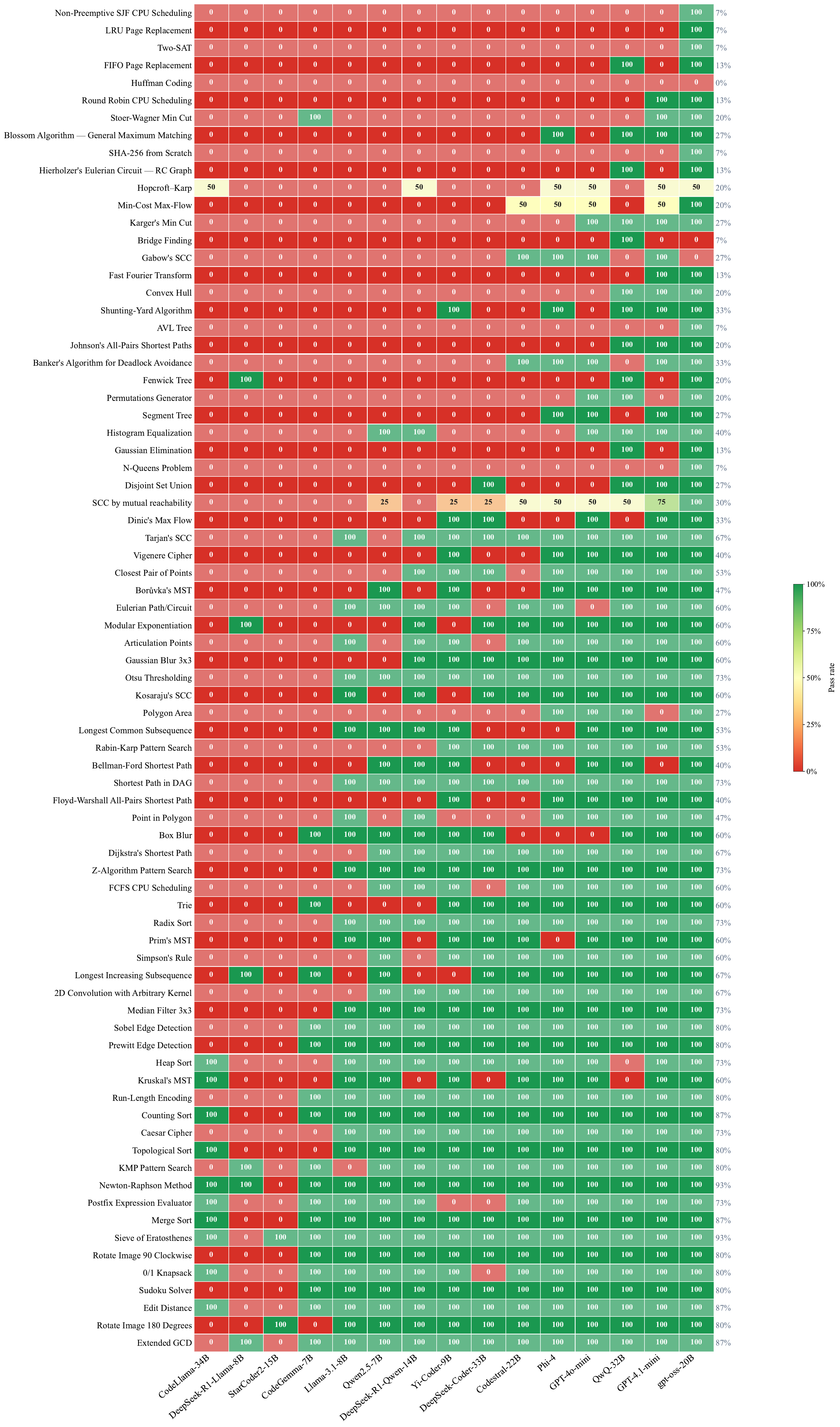}
  \caption{Algorithm difficulty heatmap for JavaScript.}
  \label{fig:heatmap-javascript}
\end{figure*}

\begin{figure*}
  \includegraphics[width=0.85\linewidth]{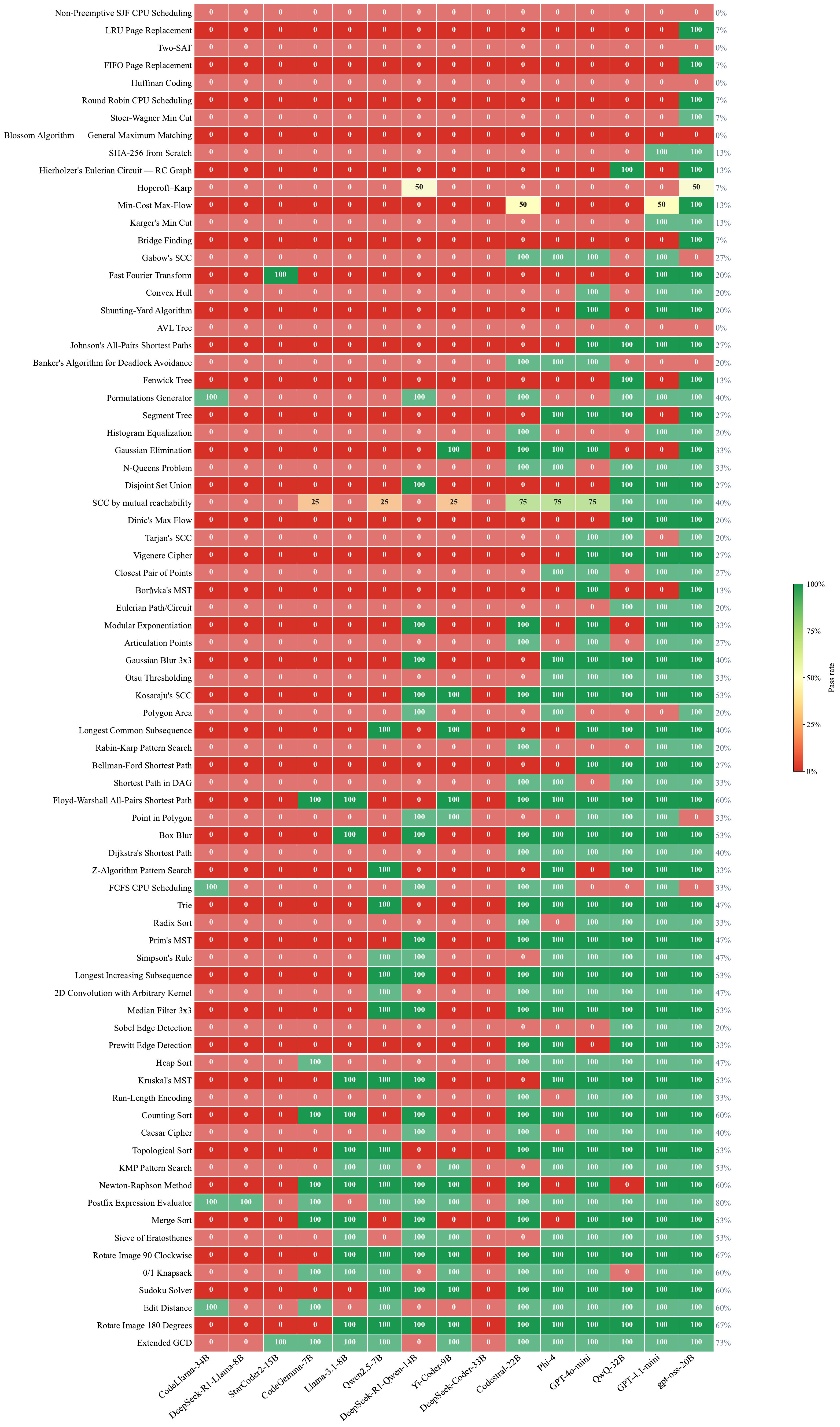}
  \caption{Algorithm difficulty heatmap for Rust.}
  \label{fig:heatmap-rust}
\end{figure*}

\begin{figure*}
  \includegraphics[width=0.85\linewidth]{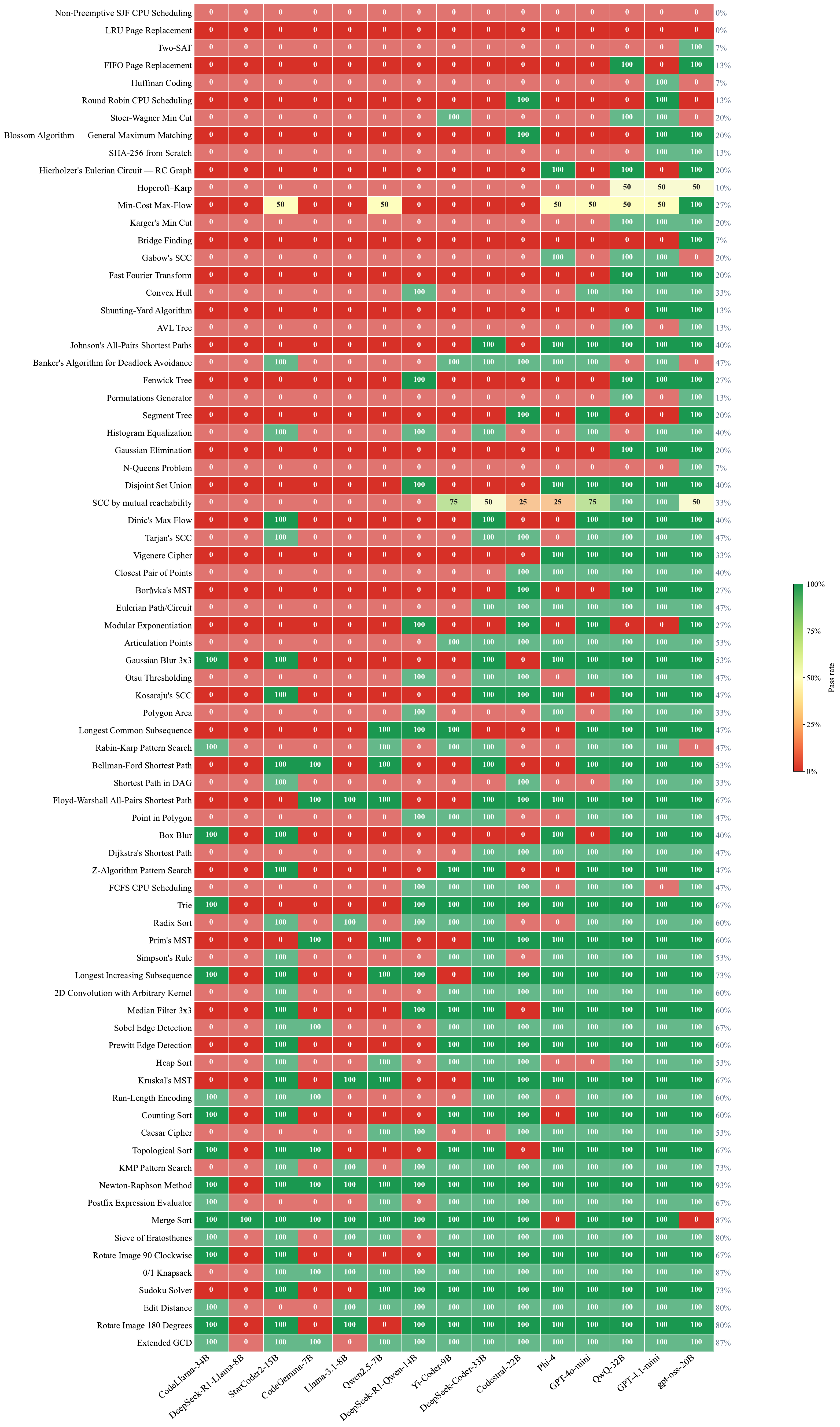}
  \caption{Algorithm difficulty heatmap for Go.}
  \label{fig:heatmap-go}
\end{figure*}

Figures~\ref{fig:heatmap-python}--\ref{fig:heatmap-go} show per-language pass-rate heatmaps for all 77 algorithms in \datasetname{} across 15 zero-shot models.
Each cell reports the fraction of trials in which the model's output matched the ground-truth expected value, averaged over all input-representation variants available for that (algorithm, language) pair.
Grey cells indicate that no predictions were collected for that combination.
Rows are sorted by global mean pass rate (averaged over all languages and all models), so the hardest algorithms appear at the top and the easiest at the bottom; this ordering is held \emph{fixed} across all seven figures to facilitate cross-language comparison.
Columns are sorted by each model's mean pass rate within the given language (weakest left, strongest right), so column order may vary between figures. The right-hand axis shows each algorithm's mean pass rate across all models for that language.

\section{Scale Experiment: A Behavioral Probe for Retrieval vs.\ Synthesis}
\label{sec:scale-experiment}
 
For each of 10 algorithms (kruskal, prim, bridges, articulation points, boruvka, dijkstra, bellman-ford, karger's min cut, Stoer--Wagner, blossom), we sample $G(n,p)$ graphs at 5 algorithm-specific values of $n$ (from $n{=}50$ for the $O(V^3)$/cut algorithms up to $n{=}8000$ for near-linear ones), 8 samples per $(algo, n)$, with fresh \texttt{networkx} ground truth per sample. We run every model that already passed \datasetname{}'s fixed-size correctness gate for that algorithm (93 of 156 candidate pairs), in isolated, resource-capped subprocesses. For each (algorithm, model), we fit a power law $\text{wall\_time} \propto n^k$ and compare the fitted exponent $k$ to the textbook exponent (e.g.\ $O(E) \to O(n^2)$ in our dense regime, where $E \sim n^2$), allowing a tolerance of $\pm0.4$. A second \emph{sparse} regime ($p(n) = 3\ln(n)/n$, $E \sim n \log n$) covers the 7 algorithms with $E$-dependent complexity: under dense sampling, a correct $O(E \log V)$ implementation and a hidden $O(V^2)$ one differ by only a log factor, but under sparse sampling they differ by a full exponent, making the two easy to tell apart.

Across both plots, the fitted runtime-vs-n curves track the theoretical (textbook) complexity slope rather than diverging from it. Most models are reproducing the right asymptotic shape, not just the right output.

\begin{figure*}[t]
    \centering
    \includegraphics[width=0.6\linewidth]{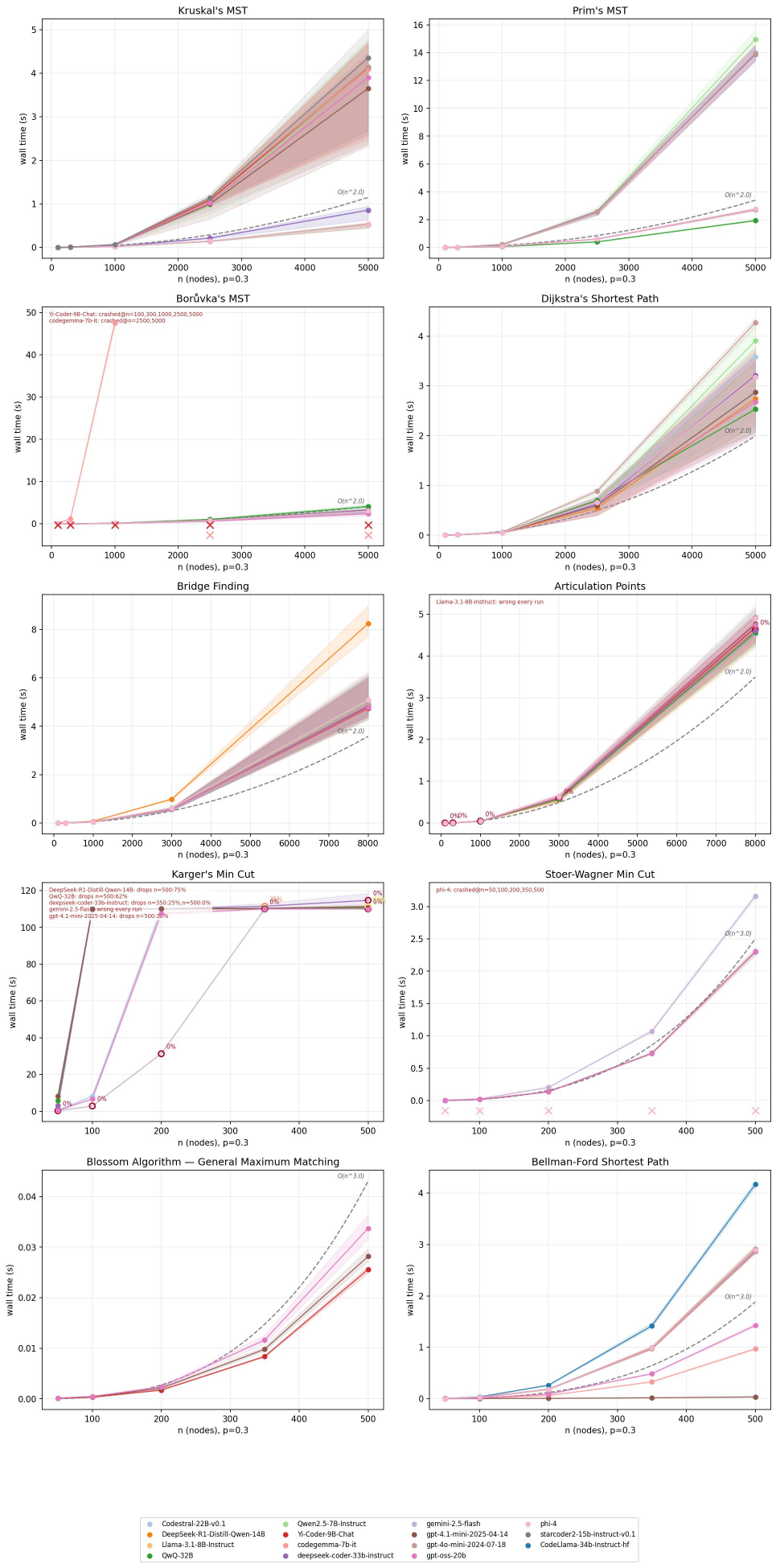}
    \caption{Dense regime ($p{=}0.3$ fixed, $E \sim n^2$): wall time vs.\ graph size $n$, one panel per algorithm and one line per model. Solid markers indicate 100\% correctness at that $n$ over 8 samples, and hollow markers (colored red-to-green) indicate a partial correct rate, and an `x' indicates every sample crashed or timed out. The dashed gray curve in each panel is the textbook Big-O reference (fit at the algorithm's expected exponent).}
    \label{fig:scale-dense}
\end{figure*}

\begin{figure*}[t]
    \centering
    \includegraphics[width=0.85\linewidth]{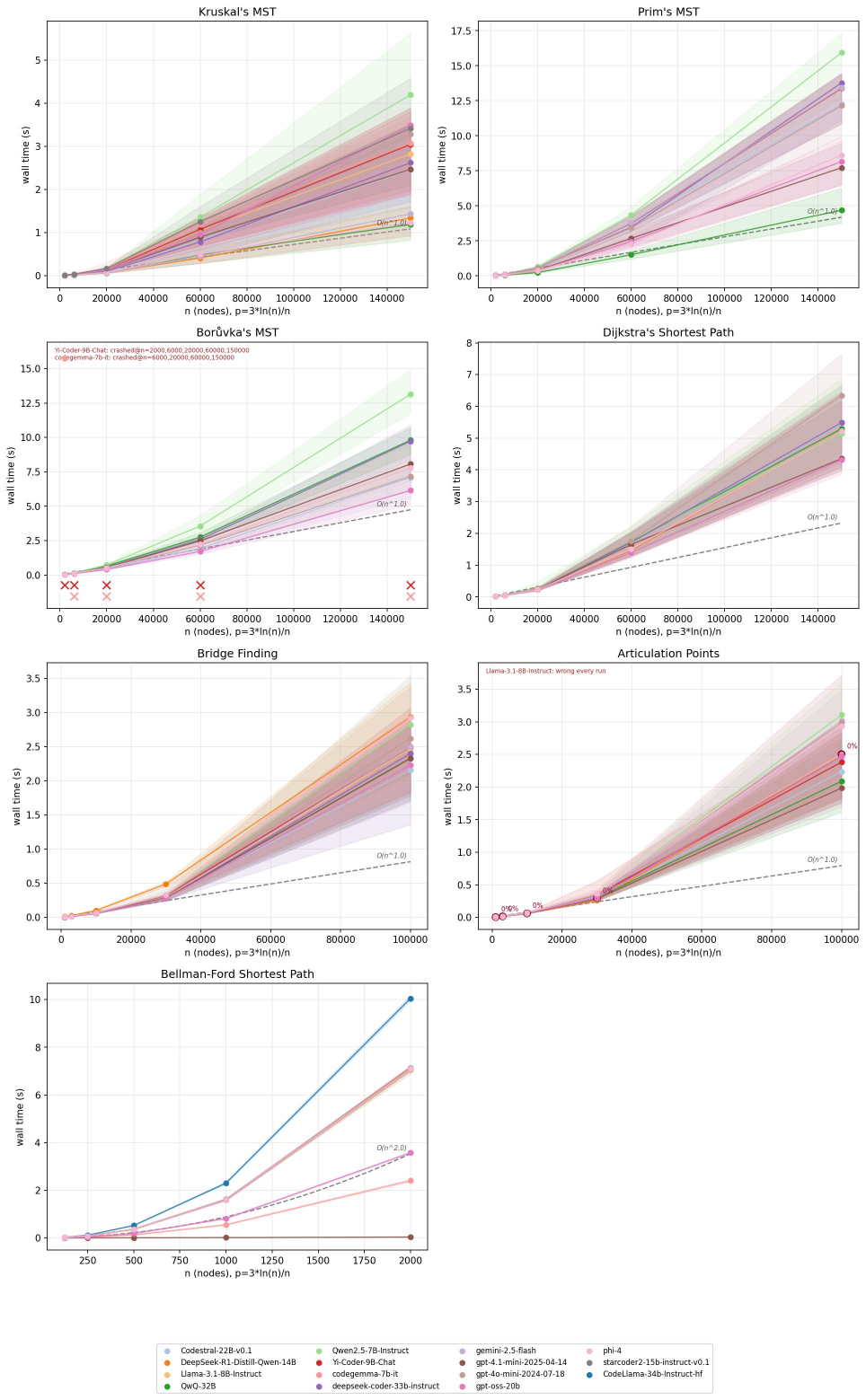}
    \caption{Sparse regime ($p(n) = 3\ln(n)/n$, $E \sim n\log n$), for the 7 algorithms whose textbook complexity is $E$-dependent. This regime separates correct $O(E \log V)$ implementations (fitted $k \approx 1$) from hidden $O(V^2)$ ones (fitted $k \approx 2$), which the dense regime alone cannot cleanly distinguish.}
    \label{fig:scale-sparse}
\end{figure*}

\section{Name/Spec Ablation for Retrieval vs.\ Synthesis}

A second, independent probe of retrieval vs.\ synthesis, complementing the scale experiment (Appendix~\ref{sec:scale-experiment}): instead of scaling input size, we manipulate the prompt itself, crossing whether the algorithm's real name is given against whether the given spec is behaviorally correct. Six novel algorithms with fabricated names and planted, unseen identifier fingerprints (Pr\"ufer sequence encoding, $k$-core decomposition, transitive reduction, tree center, bridge-tree construction, tree diameter via double-BFS) rule out memorization as an explanation for any success.

\begin{figure*}[t]
    \centering
    \includegraphics[width=1.\linewidth]{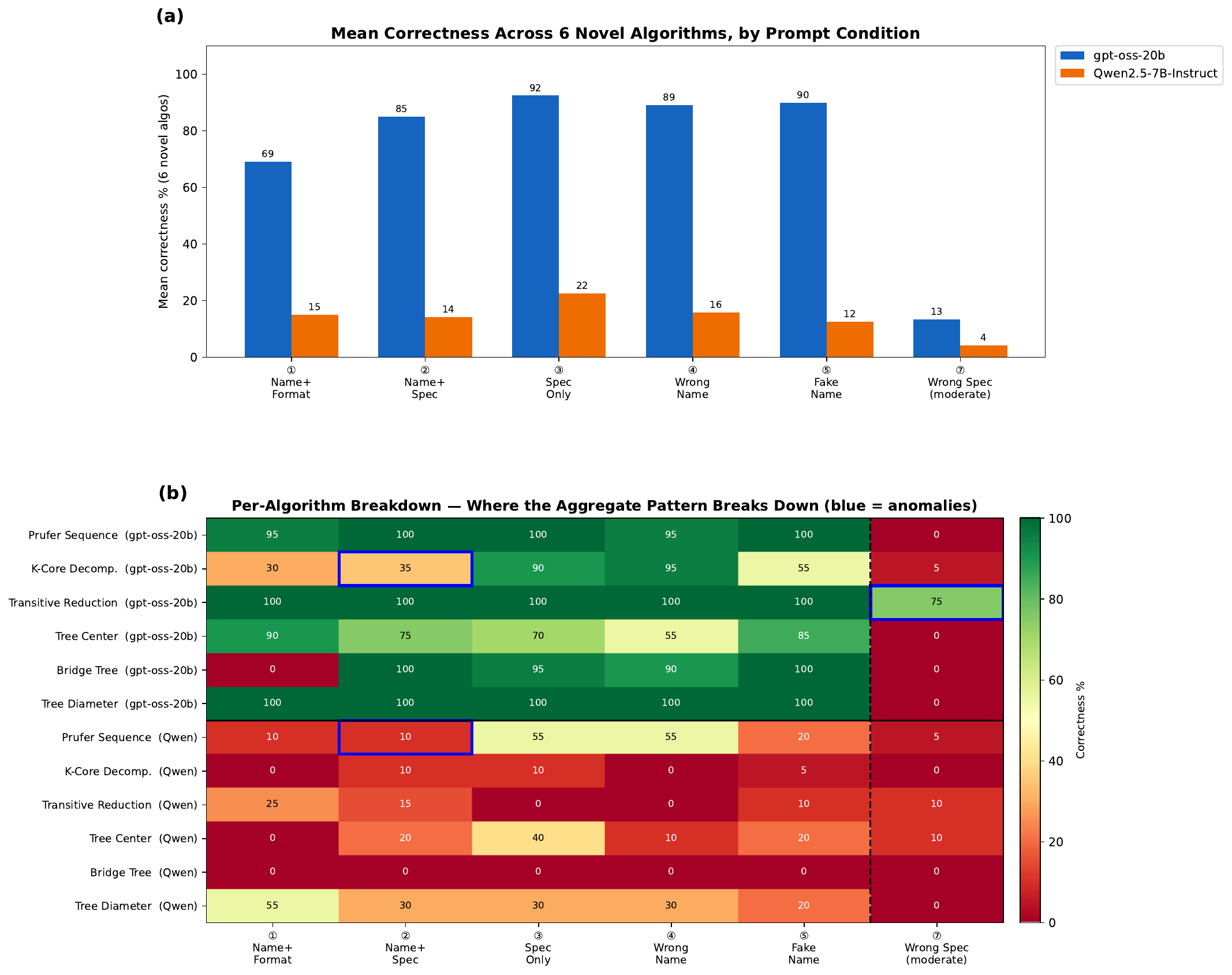}
    \caption{\textbf{(a)} Mean execution-correctness across six novel algorithms, by condition, for gpt-oss-20B and Qwen2.5-7B-Instruct (20 samples/cell). gpt-oss-20B stays flat (85--92\%) regardless of which name accompanies a correct spec, then collapses once the spec is wrong; Qwen2.5-7B-Instruct shows the same shape at a lower level. \textbf{(b)} The same conditions per algorithm (6 algorithms $\times$ 2 models). Blue-outlined cells: K-Core Decomposition and Pr\"ufer Sequence Encoding score \emph{lower} with the correct name than without it, and gpt-oss-20B's Transitive Reduction scores 75\% under moderate corruption versus 0--10\% elsewhere in that column.}
    \label{fig:option-a}
\end{figure*}

6 known \datasetname{} algorithms are tested with vs.\ without the real name (spec always correct). The 6 novel algorithms are tested under 6 conditions in Table~\ref{tab:option-a-conditions}, crossing name-correctness (absent, correct, wrong, fabricated) against spec-correctness (correct, or corrupted). So, low correctness under a corrupted spec means the model complied with the wrong instruction rather than falling back on the true algorithm. Figure~\ref{fig:option-a}a: gpt-oss-20B is flat (85--92\%) across every correct-spec condition regardless of the accompanying name, then falls to 13\% under moderate spec corruption and 0\% under severe corruption on all six algorithms, no exceptions. Qwen2.5-7B-Instruct shows the same shape at a uniformly lower level (12--22\%, falling to 4\%).

\begin{table}[htbp]
\centering
\resizebox{\linewidth}{!}{%
\begin{tabular}{lll}
\toprule
\textbf{Cond.} & \textbf{Name given} & \textbf{Spec given} \\
\midrule
(1) Name+Format   & correct     & none (signature only) \\
(2) Name+Spec     & correct     & correct \\
(3) Spec Only     & none        & correct \\
(4) Wrong Name    & wrong (real) & correct \\
(5) Fake Name     & fabricated  & correct \\
(6) Wrong Spec (moderate) & correct & one decision rule swapped \\
\bottomrule
\end{tabular}
}
\caption{The 6 conditions applied to the algorithms. All conditions are scored by execution against the ground truth.}
\label{tab:option-a-conditions}
\end{table}

\section{Algorithmic-Fidelity Judging: Spec vs.\ Code on External Benchmarks}
\label{sec:reverse-engineering}

It checks whether already-generated code on external benchmarks (MBPP, HumanEval, LiveCodeBench) implements the algorithm its prompt asked for. These benchmarks lack a single canonical algorithm name per problem, so a judge instead categorizes the general \emph{technique/algorithm} used.

A judge assigns each generation one category from a closed list (18 categories for MBPP/HumanEval, 24 for LiveCodeBench). The same judge separately predicts the expected category from the problem statement alone, standing in for a missing ground-truth label and joining spec-predicted against code-observed category per problem gives a fidelity signal. Two models (gemini-2.5-flash, gpt-4.1-mini) are judged on 150 MBPP/HumanEval and 75 LiveCodeBench problems.

We report the Pair match rate as a proxy for algorithmic fidelity, which is the fraction of (problem, model) pairs where the code-judged category equals the spec-predicted category.

The pair match rate is high on MBPP/HumanEval (80.6\%, 358/444) and lower on LiveCodeBench (61.0\%, 114/187). A second judge (GPT-4.1, vs.\ the original gemini-pro-latest) shows a clean self-preference effect: gemini-2.5-flash scores best under the Gemini judge (82.2\% MBPP, 77.5\% LiveCodeBench) but worst under GPT-4.1 (69.9\%, 53.7\%), while gpt-4.1-mini scores best under both.

\begin{thebibliography}{27}
\providecommand{\natexlab}[1]{#1}

\bibitem[{Austin et~al.(2021{\natexlab{a}})Austin, Odena, Nye, Bosma, Michalewski, Dohan, Jiang, Cai, Terry, Le, and Sutton}]{austin2021programsynthesislargelanguage}
Jacob Austin, Augustus Odena, Maxwell Nye, Maarten Bosma, Henryk Michalewski, David Dohan, Ellen Jiang, Carrie Cai, Michael Terry, Quoc Le, and Charles Sutton. 2021{\natexlab{a}}.
\newblock \href {https://arxiv.org/abs/2108.07732} {Program synthesis with large language models}.
\newblock \emph{Preprint}, arXiv:2108.07732.

\bibitem[{Austin et~al.(2021{\natexlab{b}})Austin, Odena, Nye, Bosma, Michalewski, Dohan, Jiang, Cai, Terry, Le, and Sutton}]{DBLP:journals/corr/abs-2108-07732}
Jacob Austin, Augustus Odena, Maxwell~I. Nye, Maarten Bosma, Henryk Michalewski, David Dohan, Ellen Jiang, Carrie~J. Cai, Michael Terry, Quoc~V. Le, and Charles Sutton. 2021{\natexlab{b}}.
\newblock \href {https://arxiv.org/abs/2108.07732} {Program synthesis with large language models}.
\newblock \emph{CoRR}, abs/2108.07732.

\bibitem[{Carlini et~al.(2023)Carlini, Ippolito, Jagielski, Lee, Tramer, and Zhang}]{carlini2023quantifying}
Nicholas Carlini, Daphne Ippolito, Matthew Jagielski, Katherine Lee, Florian Tramer, and Chiyuan Zhang. 2023.
\newblock \href {https://openreview.net/forum?id=TatRHT_1cK} {Quantifying memorization across neural language models}.
\newblock In \emph{The Eleventh International Conference on Learning Representations}.

\bibitem[{Cemri et~al.(2025)Cemri, Pan, Yang, Agrawal, Chopra, Tiwari, Keutzer, Parameswaran, Klein, Ramchandran, Zaharia, Gonzalez, and Stoica}]{NEURIPS2025_b1041e52}
Mert Cemri, Melissa~Z Pan, Shuyi Yang, Lakshya~A Agrawal, Bhavya Chopra, Rishabh Tiwari, Kurt Keutzer, Aditya Parameswaran, Dan Klein, Kannan Ramchandran, Matei~A Zaharia, Joseph Gonzalez, and Ion Stoica. 2025.
\newblock \href {https://proceedings.neurips.cc/paper_files/paper/2025/file/b1041e52d3be19f0a9bc491657488e4a-Paper-Datasets_and_Benchmarks_Track.pdf} {Why do multi-agent llm systems fail?}
\newblock In \emph{Advances in Neural Information Processing Systems}, volume~38. Curran Associates, Inc.

\bibitem[{Chen et~al.(2021)Chen, Tworek, Jun, Yuan, de~Oliveira~Pinto, Kaplan, Edwards, Burda, Joseph, Brockman, Ray, Puri, Krueger, Petrov, Khlaaf, Sastry, Mishkin, Chan, Gray, Ryder, Pavlov, Power, Kaiser, Bavarian, Winter, Tillet, Such, Cummings, Plappert, Chantzis, Barnes, Herbert-Voss, Guss, Nichol, Paino, Tezak, Tang, Babuschkin, Balaji, Jain, Saunders, Hesse, Carr, Leike, Achiam, Misra, Morikawa, Radford, Knight, Brundage, Murati, Mayer, Welinder, McGrew, Amodei, McCandlish, Sutskever, and Zaremba}]{chen2021codex}
Mark Chen, Jerry Tworek, Heewoo Jun, Qiming Yuan, Henrique~Ponde de~Oliveira~Pinto, Jared Kaplan, Harri Edwards, Yuri Burda, Nicholas Joseph, Greg Brockman, Alex Ray, Raul Puri, Gretchen Krueger, Michael Petrov, Heidy Khlaaf, Girish Sastry, Pamela Mishkin, Brooke Chan, Scott Gray, and 39 others. 2021.
\newblock \href {https://arxiv.org/abs/2107.03374} {Evaluating large language models trained on code}.

\bibitem[{Gao et~al.(2020)Gao, Biderman, Black, Golding, Hoppe, Foster, Phang, He, Thite, Nabeshima, Presser, and Leahy}]{gao2020pile800gbdatasetdiverse}
Leo Gao, Stella Biderman, Sid Black, Laurence Golding, Travis Hoppe, Charles Foster, Jason Phang, Horace He, Anish Thite, Noa Nabeshima, Shawn Presser, and Connor Leahy. 2020.
\newblock \href {https://arxiv.org/abs/2101.00027} {The pile: An 800gb dataset of diverse text for language modeling}.
\newblock \emph{Preprint}, arXiv:2101.00027.

\bibitem[{Gekhman et~al.(2026)Gekhman, Aharoni, Ofek, Geva, Reichart, and Herzig}]{gekhman2026thinkingrecallreasoningunlocks}
Zorik Gekhman, Roee Aharoni, Eran Ofek, Mor Geva, Roi Reichart, and Jonathan Herzig. 2026.
\newblock \href {https://arxiv.org/abs/2603.09906} {Thinking to recall: How reasoning unlocks parametric knowledge in llms}.
\newblock \emph{Preprint}, arXiv:2603.09906.

\bibitem[{Grattafiori et~al.(2024)Grattafiori, Dubey, Jauhri, Pandey, Kadian, Al-Dahle, Letman, Mathur, Schelten, Vaughan, Yang, Fan, Goyal, Hartshorn, Yang, Mitra, Sravankumar, Korenev, Hinsvark, Rao, Zhang, Rodriguez, Gregerson, Spataru, Roziere, Biron, Tang, Chern, Caucheteux, Nayak, Bi, Marra, McConnell, Keller, Touret, Wu, Wong, Ferrer, Nikolaidis, Allonsius, Song, Pintz, Livshits, Wyatt, Esiobu, Choudhary, Mahajan, Garcia-Olano, Perino, Hupkes, Lakomkin, AlBadawy, Lobanova, Dinan, Smith, Radenovic, Guzmán, Zhang, Synnaeve, Lee, Anderson, Thattai, Nail, Mialon, Pang, Cucurell, Nguyen, Korevaar, Xu, Touvron, Zarov, Ibarra, Kloumann, Misra, Evtimov, Zhang, Copet, Lee, Geffert, Vranes, Park, Mahadeokar, Shah, van~der Linde, Billock, Hong, Lee, Fu, Chi, Huang, Liu, Wang, Yu, Bitton, Spisak, Park, Rocca, Johnstun, Saxe, Jia, Alwala, Prasad, Upasani, Plawiak, Li, Heafield, Stone, El-Arini, Iyer, Malik, Chiu, Bhalla, Lakhotia, Rantala-Yeary, van~der Maaten, Chen, Tan, Jenkins, Martin, Madaan, Malo, Blecher,
  Landzaat, de~Oliveira, Muzzi, Pasupuleti, Singh, Paluri, Kardas, Tsimpoukelli, Oldham, Rita, Pavlova, Kambadur, Lewis, Si, Singh, Hassan, Goyal, Torabi, Bashlykov, Bogoychev, Chatterji, Zhang, Duchenne, Çelebi, Alrassy, Zhang, Li, Vasic, Weng, Bhargava, Dubal, Krishnan, Koura, Xu, He, Dong, Srinivasan, Ganapathy, Calderer, Cabral, Stojnic, Raileanu, Maheswari, Girdhar, Patel, Sauvestre, Polidoro, Sumbaly, Taylor, Silva, Hou, Wang, Hosseini, Chennabasappa, Singh, Bell, Kim, Edunov, Nie, Narang, Raparthy, Shen, Wan, Bhosale, Zhang, Vandenhende, Batra, Whitman, Sootla, Collot, Gururangan, Borodinsky, Herman, Fowler, Sheasha, Georgiou, Scialom, Speckbacher, Mihaylov, Xiao, Karn, Goswami, Gupta, Ramanathan, Kerkez, Gonguet, Do, Vogeti, Albiero, Petrovic, Chu, Xiong, Fu, Meers, Martinet, Wang, Wang, Tan, Xia, Xie, Jia, Wang, Goldschlag, Gaur, Babaei, Wen, Song, Zhang, Li, Mao, Coudert, Yan, Chen, Papakipos, Singh, Srivastava, Jain, Kelsey, Shajnfeld, Gangidi, Victoria, Goldstand, Menon, Sharma, Boesenberg,
  Baevski, Feinstein, Kallet, Sangani, Teo, Yunus, Lupu, Alvarado, Caples, Gu, Ho, Poulton, Ryan, Ramchandani, Dong, Franco, Goyal, Saraf, Chowdhury, Gabriel, Bharambe, Eisenman, Yazdan, James, Maurer, Leonhardi, Huang, Loyd, Paola, Paranjape, Liu, Wu, Ni, Hancock, Wasti, Spence, Stojkovic, Gamido, Montalvo, Parker, Burton, Mejia, Liu, Wang, Kim, Zhou, Hu, Chu, Cai, Tindal, Feichtenhofer, Gao, Civin, Beaty, Kreymer, Li, Adkins, Xu, Testuggine, David, Parikh, Liskovich, Foss, Wang, Le, Holland, Dowling, Jamil, Montgomery, Presani, Hahn, Wood, Le, Brinkman, Arcaute, Dunbar, Smothers, Sun, Kreuk, Tian, Kokkinos, Ozgenel, Caggioni, Kanayet, Seide, Florez, Schwarz, Badeer, Swee, Halpern, Herman, Sizov, Guangyi, Zhang, Lakshminarayanan, Inan, Shojanazeri, Zou, Wang, Zha, Habeeb, Rudolph, Suk, Aspegren, Goldman, Zhan, Damlaj, Molybog, Tufanov, Leontiadis, Veliche, Gat, Weissman, Geboski, Kohli, Lam, Asher, Gaya, Marcus, Tang, Chan, Zhen, Reizenstein, Teboul, Zhong, Jin, Yang, Cummings, Carvill, Shepard, McPhie,
  Torres, Ginsburg, Wang, Wu, U, Saxena, Khandelwal, Zand, Matosich, Veeraraghavan, Michelena, Li, Jagadeesh, Huang, Chawla, Huang, Chen, Garg, A, Silva, Bell, Zhang, Guo, Yu, Moshkovich, Wehrstedt, Khabsa, Avalani, Bhatt, Mankus, Hasson, Lennie, Reso, Groshev, Naumov, Lathi, Keneally, Liu, Seltzer, Valko, Restrepo, Patel, Vyatskov, Samvelyan, Clark, Macey, Wang, Hermoso, Metanat, Rastegari, Bansal, Santhanam, Parks, White, Bawa, Singhal, Egebo, Usunier, Mehta, Laptev, Dong, Cheng, Chernoguz, Hart, Salpekar, Kalinli, Kent, Parekh, Saab, Balaji, Rittner, Bontrager, Roux, Dollar, Zvyagina, Ratanchandani, Yuvraj, Liang, Alao, Rodriguez, Ayub, Murthy, Nayani, Mitra, Parthasarathy, Li, Hogan, Battey, Wang, Howes, Rinott, Mehta, Siby, Bondu, Datta, Chugh, Hunt, Dhillon, Sidorov, Pan, Mahajan, Verma, Yamamoto, Ramaswamy, Lindsay, Lindsay, Feng, Lin, Zha, Patil, Shankar, Zhang, Zhang, Wang, Agarwal, Sajuyigbe, Chintala, Max, Chen, Kehoe, Satterfield, Govindaprasad, Gupta, Deng, Cho, Virk, Subramanian, Choudhury,
  Goldman, Remez, Glaser, Best, Koehler, Robinson, Li, Zhang, Matthews, Chou, Shaked, Vontimitta, Ajayi, Montanez, Mohan, Kumar, Mangla, Ionescu, Poenaru, Mihailescu, Ivanov, Li, Wang, Jiang, Bouaziz, Constable, Tang, Wu, Wang, Wu, Gao, Kleinman, Chen, Hu, Jia, Qi, Li, Zhang, Zhang, Adi, Nam, Yu, Wang, Zhao, Hao, Qian, Li, He, Rait, DeVito, Rosnbrick, Wen, Yang, Zhao, and Ma}]{grattafiori2024llama3herdmodels}
Aaron Grattafiori, Abhimanyu Dubey, Abhinav Jauhri, Abhinav Pandey, Abhishek Kadian, Ahmad Al-Dahle, Aiesha Letman, Akhil Mathur, Alan Schelten, Alex Vaughan, Amy Yang, Angela Fan, Anirudh Goyal, Anthony Hartshorn, Aobo Yang, Archi Mitra, Archie Sravankumar, Artem Korenev, Arthur Hinsvark, and 542 others. 2024.
\newblock \href {https://arxiv.org/abs/2407.21783} {The llama 3 herd of models}.
\newblock \emph{Preprint}, arXiv:2407.21783.

\bibitem[{Guo et~al.(2024)Guo, Zhu, Yang, Xie, Dong, Zhang, Chen, Bi, Wu, Li, Luo, Xiong, and Liang}]{guo2024deepseekcoderlargelanguagemodel}
Daya Guo, Qihao Zhu, Dejian Yang, Zhenda Xie, Kai Dong, Wentao Zhang, Guanting Chen, Xiao Bi, Y.~Wu, Y.~K. Li, Fuli Luo, Yingfei Xiong, and Wenfeng Liang. 2024.
\newblock \href {https://arxiv.org/abs/2401.14196} {Deepseek-coder: When the large language model meets programming -- the rise of code intelligence}.
\newblock \emph{Preprint}, arXiv:2401.14196.

\bibitem[{Hendrycks et~al.(2021)Hendrycks, Basart, Kadavath, Mazeika, Arora, Guo, Burns, Puranik, He, Song, and Steinhardt}]{hendrycksapps2021}
Dan Hendrycks, Steven Basart, Saurav Kadavath, Mantas Mazeika, Akul Arora, Ethan Guo, Collin Burns, Samir Puranik, Horace He, Dawn Song, and Jacob Steinhardt. 2021.
\newblock Measuring coding challenge competence with apps.
\newblock \emph{NeurIPS}.

\bibitem[{Huang et~al.(2024)Huang, Yang, and Potts}]{huang-etal-2024-demystifying}
Jing Huang, Diyi Yang, and Christopher Potts. 2024.
\newblock \href {https://doi.org/10.18653/v1/2024.emnlp-main.598} {Demystifying verbatim memorization in large language models}.
\newblock In \emph{Proceedings of the 2024 Conference on Empirical Methods in Natural Language Processing}, pages 10711--10732, Miami, Florida, USA. Association for Computational Linguistics.

\bibitem[{Jain et~al.(2024)Jain, Han, Gu, Li, Yan, Zhang, Wang, Solar-Lezama, Sen, and Stoica}]{jain2024livecodebenchholisticcontaminationfree}
Naman Jain, King Han, Alex Gu, Wen-Ding Li, Fanjia Yan, Tianjun Zhang, Sida Wang, Armando Solar-Lezama, Koushik Sen, and Ion Stoica. 2024.
\newblock \href {https://arxiv.org/abs/2403.07974} {Livecodebench: Holistic and contamination free evaluation of large language models for code}.
\newblock \emph{Preprint}, arXiv:2403.07974.

\bibitem[{Kocetkov et~al.(2022)Kocetkov, Li, Allal, Li, Mou, Ferrandis, Jernite, Mitchell, Hughes, Wolf, Bahdanau, von Werra, and de~Vries}]{kocetkov2022stack3tbpermissively}
Denis Kocetkov, Raymond Li, Loubna~Ben Allal, Jia Li, Chenghao Mou, Carlos~Muñoz Ferrandis, Yacine Jernite, Margaret Mitchell, Sean Hughes, Thomas Wolf, Dzmitry Bahdanau, Leandro von Werra, and Harm de~Vries. 2022.
\newblock \href {https://arxiv.org/abs/2211.15533} {The stack: 3 tb of permissively licensed source code}.
\newblock \emph{Preprint}, arXiv:2211.15533.

\bibitem[{Lee et~al.(2025)Lee, Chon, Jang, Lee, and Yu}]{lee-etal-2025-diversely}
Seonghyeon Lee, HeeJae Chon, Joonwon Jang, Dongha Lee, and Hwanjo Yu. 2025.
\newblock \href {https://doi.org/10.18653/v1/2025.findings-emnlp.10} {How diversely can language models solve problems? exploring the algorithmic diversity of model-generated code}.
\newblock In \emph{Findings of the Association for Computational Linguistics: EMNLP 2025}, pages 152--167, Suzhou, China. Association for Computational Linguistics.

\bibitem[{Liu et~al.(2023)Liu, Xia, Wang, and Zhang}]{evalplus}
Jiawei Liu, Chunqiu~Steven Xia, Yuyao Wang, and Lingming Zhang. 2023.
\newblock \href {https://openreview.net/forum?id=1qvx610Cu7} {Is your code generated by chat{GPT} really correct? rigorous evaluation of large language models for code generation}.
\newblock In \emph{Thirty-seventh Conference on Neural Information Processing Systems}.

\bibitem[{Lozhkov et~al.(2024)Lozhkov, Li, Allal, Cassano, Lamy-Poirier, Tazi, Tang, Pykhtar, Liu, Wei, Liu, Tian, Kocetkov, Zucker, Belkada, Wang, Liu, Abulkhanov, Paul, Li, Li, Risdal, Li, Zhu, Zhuo, Zheltonozhskii, Dade, Yu, Krauß, Jain, Su, He, Dey, Abati, Chai, Muennighoff, Tang, Oblokulov, Akiki, Marone, Mou, Mishra, Gu, Hui, Dao, Zebaze, Dehaene, Patry, Xu, McAuley, Hu, Scholak, Paquet, Robinson, Anderson, Chapados, Patwary, Tajbakhsh, Jernite, Ferrandis, Zhang, Hughes, Wolf, Guha, von Werra, and de~Vries}]{lozhkov2024starcoder2stackv2}
Anton Lozhkov, Raymond Li, Loubna~Ben Allal, Federico Cassano, Joel Lamy-Poirier, Nouamane Tazi, Ao~Tang, Dmytro Pykhtar, Jiawei Liu, Yuxiang Wei, Tianyang Liu, Max Tian, Denis Kocetkov, Arthur Zucker, Younes Belkada, Zijian Wang, Qian Liu, Dmitry Abulkhanov, Indraneil Paul, and 47 others. 2024.
\newblock \href {https://arxiv.org/abs/2402.19173} {Starcoder 2 and the stack v2: The next generation}.
\newblock \emph{Preprint}, arXiv:2402.19173.

\bibitem[{Matton et~al.(2024)Matton, Sherborne, Aumiller, Tommasone, Alizadeh, He, Ma, Voisin, Gilsenan-McMahon, and Gall{\'e}}]{matton-etal-2024-leakage}
Alexandre Matton, Tom Sherborne, Dennis Aumiller, Elena Tommasone, Milad Alizadeh, Jingyi He, Raymond Ma, Maxime Voisin, Ellen Gilsenan-McMahon, and Matthias Gall{\'e}. 2024.
\newblock \href {https://doi.org/10.18653/v1/2024.findings-emnlp.772} {On leakage of code generation evaluation datasets}.
\newblock In \emph{Findings of the Association for Computational Linguistics: EMNLP 2024}, pages 13215--13223, Miami, Florida, USA. Association for Computational Linguistics.

\bibitem[{Maveli et~al.(2026)Maveli, Vergari, and Cohen}]{maveli2026llmscompressanddecompress}
Nickil Maveli, Antonio Vergari, and Shay~B. Cohen. 2026.
\newblock \href {https://arxiv.org/abs/2601.13398} {Can llms compress (and decompress)? evaluating code understanding and execution via invertibility}.
\newblock \emph{Preprint}, arXiv:2601.13398.

\bibitem[{Petroni et~al.(2019)Petroni, Rockt{\"a}schel, Riedel, Lewis, Bakhtin, Wu, and Miller}]{petroni-etal-2019-language}
Fabio Petroni, Tim Rockt{\"a}schel, Sebastian Riedel, Patrick Lewis, Anton Bakhtin, Yuxiang Wu, and Alexander Miller. 2019.
\newblock \href {https://doi.org/10.18653/v1/D19-1250} {Language models as knowledge bases?}
\newblock In \emph{Proceedings of the 2019 Conference on Empirical Methods in Natural Language Processing and the 9th International Joint Conference on Natural Language Processing (EMNLP-IJCNLP)}, pages 2463--2473, Hong Kong, China. Association for Computational Linguistics.

\bibitem[{Riddell et~al.(2024)Riddell, Ni, and Cohan}]{riddell-etal-2024-quantifying}
Martin Riddell, Ansong Ni, and Arman Cohan. 2024.
\newblock \href {https://doi.org/10.18653/v1/2024.acl-long.761} {Quantifying contamination in evaluating code generation capabilities of language models}.
\newblock In \emph{Proceedings of the 62nd Annual Meeting of the Association for Computational Linguistics (Volume 1: Long Papers)}, pages 14116--14137, Bangkok, Thailand. Association for Computational Linguistics.

\bibitem[{Roberts et~al.(2020)Roberts, Raffel, and Shazeer}]{roberts-etal-2020-much}
Adam Roberts, Colin Raffel, and Noam Shazeer. 2020.
\newblock \href {https://doi.org/10.18653/v1/2020.emnlp-main.437} {How much knowledge can you pack into the parameters of a language model?}
\newblock In \emph{Proceedings of the 2020 Conference on Empirical Methods in Natural Language Processing (EMNLP)}, pages 5418--5426, Online. Association for Computational Linguistics.

\bibitem[{Sapkota et~al.(2025)Sapkota, Roumeliotis, and Karkee}]{sapkota2025vibecodingvsagentic}
Ranjan Sapkota, Konstantinos~I. Roumeliotis, and Manoj Karkee. 2025.
\newblock \href {https://arxiv.org/abs/2505.19443} {Vibe coding vs. agentic coding: Fundamentals and practical implications of agentic ai}.
\newblock \emph{Preprint}, arXiv:2505.19443.

\bibitem[{Shao et~al.(2024)Shao, Wang, Zhu, Xu, Song, Bi, Zhang, Zhang, Li, Wu, and Guo}]{shao2024deepseekmathpushinglimitsmathematical}
Zhihong Shao, Peiyi Wang, Qihao Zhu, Runxin Xu, Junxiao Song, Xiao Bi, Haowei Zhang, Mingchuan Zhang, Y.~K. Li, Y.~Wu, and Daya Guo. 2024.
\newblock \href {https://arxiv.org/abs/2402.03300} {Deepseekmath: Pushing the limits of mathematical reasoning in open language models}.
\newblock \emph{Preprint}, arXiv:2402.03300.

\bibitem[{Soldaini et~al.(2024)Soldaini, Kinney, Bhagia, Schwenk, Atkinson, Authur, Bogin, Chandu, Dumas, Elazar, Hofmann, Jha, Kumar, Lucy, Lyu, Lambert, Magnusson, Morrison, Muennighoff, Naik, Nam, Peters, Ravichander, Richardson, Shen, Strubell, Subramani, Tafjord, Walsh, Zettlemoyer, Smith, Hajishirzi, Beltagy, Groeneveld, Dodge, and Lo}]{dolma}
Luca Soldaini, Rodney Kinney, Akshita Bhagia, Dustin Schwenk, David Atkinson, Russell Authur, Ben Bogin, Khyathi Chandu, Jennifer Dumas, Yanai Elazar, Valentin Hofmann, Ananya~Harsh Jha, Sachin Kumar, Li~Lucy, Xinxi Lyu, Nathan Lambert, Ian Magnusson, Jacob Morrison, Niklas Muennighoff, and 17 others. 2024.
\newblock \href {https://arxiv.org/abs/2402.00159} {{Dolma: An Open Corpus of Three Trillion Tokens for Language Model Pretraining Research}}.
\newblock \emph{arXiv preprint}.

\bibitem[{Touvron et~al.(2023)Touvron, Lavril, Izacard, Martinet, Lachaux, Lacroix, Rozière, Goyal, Hambro, Azhar, Rodriguez, Joulin, Grave, and Lample}]{touvron2023llamaopenefficientfoundation}
Hugo Touvron, Thibaut Lavril, Gautier Izacard, Xavier Martinet, Marie-Anne Lachaux, Timothée Lacroix, Baptiste Rozière, Naman Goyal, Eric Hambro, Faisal Azhar, Aurelien Rodriguez, Armand Joulin, Edouard Grave, and Guillaume Lample. 2023.
\newblock \href {https://arxiv.org/abs/2302.13971} {Llama: Open and efficient foundation language models}.
\newblock \emph{Preprint}, arXiv:2302.13971.

\bibitem[{Wang et~al.(2025)Wang, Asai, Yu, Xu, Xie, Neubig, and Fried}]{wang-etal-2025-coderag}
Zora~Zhiruo Wang, Akari Asai, Xinyan~Velocity Yu, Frank~F. Xu, Yiqing Xie, Graham Neubig, and Daniel Fried. 2025.
\newblock \href {https://doi.org/10.18653/v1/2025.findings-naacl.176} {{C}ode{RAG}-bench: Can retrieval augment code generation?}
\newblock In \emph{Findings of the Association for Computational Linguistics: NAACL 2025}, pages 3199--3214, Albuquerque, New Mexico. Association for Computational Linguistics.

\bibitem[{Zhu et~al.(2025)Zhu, Liu, Li, Tian, Yang, Zhang, Han, Xie, Cui, Zhang, Ma, Yu, Ramesh, Wu, Liu, Lu, Zou, and You}]{zhu2025llmagentsfaillearn}
Kunlun Zhu, Zijia Liu, Bingxuan Li, Muxin Tian, Yingxuan Yang, Jiaxun Zhang, Pengrui Han, Qipeng Xie, Fuyang Cui, Weijia Zhang, Xiaoteng Ma, Xiaodong Yu, Gowtham Ramesh, Jialian Wu, Zicheng Liu, Pan Lu, James Zou, and Jiaxuan You. 2025.
\newblock \href {https://arxiv.org/abs/2509.25370} {Where llm agents fail and how they can learn from failures}.
\newblock \emph{Preprint}, arXiv:2509.25370.

\end{thebibliography}
\end{document}